\documentclass[12pt]{article}
\usepackage{verbatim,algorithm,algpseudocode}
\usepackage[symbol]{footmisc}
\usepackage{xr}

\usepackage{graphicx}
\RequirePackage{amsthm,amsmath,amsfonts,amssymb}
\usepackage{mathrsfs}
\usepackage{algorithm}
\usepackage{algpseudocode}
\usepackage{mathtools}
\usepackage{microtype}
\usepackage{bm,booktabs,float}
\usepackage[round, authoryear]{natbib}
\usepackage{hyperref}
\usepackage{makecell}
\usepackage{multirow}
\usepackage{multicol}
\usepackage{tabularx}
\usepackage[dvipsnames]{xcolor}
\usepackage{enumitem,comment}
\usepackage{thmtools}
\usepackage{thm-restate}

\usepackage{graphicx}
\usepackage{subcaption}
\usepackage{setspace}
\allowdisplaybreaks

\numberwithin{equation}{section}
\theoremstyle{plain}

\newtheorem{assumption}{Assumption}
\newtheorem{definition}{Definition}
\newtheorem{corollary}{Corollary}[section]

\newtheorem{proposition}{Proposition}[section]
\newtheorem{remark}{Remark}

\newcommand{\br}{\boldsymbol{r}}
\newcommand{\bu}{\boldsymbol{u}}
\newcommand{\bv}{\boldsymbol{v}}
\newcommand{\bs}{\boldsymbol{s}}

\newcommand{\bbf}{\boldsymbol{f}}
\newcommand{\bx}{\boldsymbol{x}}
\newcommand{\bb}{\boldsymbol{b}}

\newcommand{\by}{\boldsymbol{y}}
\newcommand{\bz}{\boldsymbol{z}}
\newcommand{\bI}{\boldsymbol{I}}
\newcommand{\inde}{\perp\!\!\!\perp}

\newcommand{\cF}{\mathcal{F}}

\newcommand{\Rbb}{\mathbb{R}}
\newcommand{\Ebb}{\mathbb{E}}
\newcommand{\diff}{\mathop{}\!\mathrm{d}}

\newcommand{\Dkl}{D_{\text{KL}}}

\newcommand{\blind}{1}

\usepackage{hyperref}
\hypersetup{
    colorlinks=true,       
    linkcolor=blue,        
    filecolor=magenta,     
    urlcolor=cyan,         
    citecolor=blue,       
}

\begin{document}

\def\spacingset#1{\renewcommand{\baselinestretch}%
{#1}\small\normalsize}
\spacingset{1}


\if1\blind
{
  \title{Semi-Supervised Conditional Generative Learning through Stochastic Interpolation and Sufficient Representations}
  \author{Changyu Liu\thanks{Institute for Mathematics and Artificial Intelligence, Wuhan University, Wuhan, China. Email: changyuliu@whu.edu.cn},
  \ Yuling Jiao\thanks{School of Artificial Intelligence, Hubei Key Laboratory of Computational Science, Wuhan University, Wuhan, China. Email: yulingjiaomath@whu.edu.cn},
  \ and Jian Huang\thanks{Departments of Data Science and Artificial Intelligence, and Applied Mathematics, The Hong Kong Polytechnic University, Hong Kong, China. Email: j.huang@polyu.edu.hk}
  }
  \date{}
  \maketitle
 }
\fi

\if0\blind
{
  \title{Semi-Supervised Conditional Generative Learning through Stochastic Interpolation and Sufficient Representations}
  \date{}
  \maketitle
} \fi

\bigskip
\begin{abstract} 
Conditional generative modeling remains a challenging problem in semi-supervised settings where labeled data is scarce but unlabeled samples are abundant. To effectively leverage structural information embedded within the unlabeled dataset and compensate for sparse conditioning signals, we propose a semi-supervised framework combining conditional stochastic interpolation with low-dimensional latent representations. RepG decomposes generation into two stages: label-dependent latent sampling and high-dimensional reconstruction. This isolates the supervised learning of conditional dependencies to a low-dimensional space, requiring few labels while utilizing the abundant unlabeled data purely for reconstruction. Theoretically, we establish an error decomposition showing that the Kullback-Leibler divergence of RepG comprises stage-wise estimation errors and a structural bias quantified by conditional mutual information. For deep neural network estimators, we derive non-asymptotic convergence rates proving that RepG significantly improves sample complexity. By confining the supervised estimation burden to the low intrinsic dimension of the latent representation, RepG achieves a strictly faster convergence rate. Complemented by a minimax lower bound, our theoretical results demonstrate that this method effectively mitigates the curse of dimensionality inherent in direct ambient-space generative modeling.
\end{abstract}

\noindent%

\vfill

\newpage
\spacingset{1.9} 


\section{Introduction}\label{sec:intro}
Conditional generative modeling seeks to learn the underlying conditional distribution $P_{X \mid Y}$ of high-dimensional data $X \in \mathbb{R}^d$ given a specific condition $Y \in \mathbb{R}^q$. Unlike standard regression frameworks that yield only point estimates, learning the full conditional distribution enables rigorous uncertainty quantification and preserves complex structural features, such as multimodality and heteroscedasticity. While these capabilities are essential for diverse applications, ranging from image synthesis \citep{ho2021classifierfree} to inverse problems \citep{song2022solving} and scenario generation \citep{Yang2025}, they are difficult to achieve in high-dimensional spaces. Standard generative models typically approximate $P_{X \mid Y}$ by directly injecting the conditioning variable $Y$ into the high-dimensional generation process. 
However, learning the complex ambient conditional distribution $P_{X \mid Y}$ is highly data-intensive, requiring substantial labeled data. Consequently, this direct approach suffers from poor sample efficiency in semi-supervised settings, where labeled pairs $(X, Y)$ are scarce despite an abundance of unlabeled samples of $X$.

To address this difficulty, we propose a representation-based framework that factorizes the conditional generation process. Rather than modeling the conditional distributions directly in the high-dimensional ambient space, we introduce a low-dimensional latent representation $R(X) \in \mathbb{R}^k$ (where $k \ll d$) designed to capture all the essential information governing the dependence between the target data $X$ and its condition $Y$.

A naive representation-based approach directly decomposes the conditional generation process into two stages: first generating a latent  representation $R$ from the specific condition $Y$, and subsequently generating the  high-dimensional target $X$ conditioned on both $R$ and $Y$. However,  $Y$ is strictly required in both stages of this general formulation; therefore, training such a model  relies entirely on  labeled data $(X, Y)$. This persistent dependence on explicit labels critically prevents the framework from leveraging large repositories of unlabeled data.

We overcome this limitation by using a \textit{sufficient representation} \citep{Li1991, Deepnonlinear}. A representation is sufficient if it captures all the information necessary to link the target data to its condition. In other words, once this representation is known, the final data generation process becomes independent of the original condition. This property allows us to partition the generation process into two distinct, sequential stages:
\begin{enumerate}
    \item \textit{Latent sampling:} A low-dimensional step that generates the sufficient representation based on the  condition.
    \item \textit{Reconstruction:} A step that maps the generated representation into the final data space, completely bypassing the need to reference the original condition.
\end{enumerate}
By effectively circumventing the conditioning variable during the complex, high-dimensional generation stage, this structural factorization  enables scalable and efficient training using primarily unlabeled data. We refer to our method as RepG to emphasize the crucial role of data representation in the generation process.

This structural separation is particularly advantageous for semi-supervised learning. The statistically demanding high-dimensional reconstruction stage can be trained entirely using the abundant unlabeled data. Concurrently, the scarce labeled data are  reserved for learning the much simpler, low-dimensional mapping between the condition and the latent space. Consequently, this separation  mitigates the statistical curse of dimensionality inherent in direct conditional generative modeling. Even when the underlying assumption of perfect sufficiency holds only approximately in practice, this modular decomposition remains a highly effective structural prior. As demonstrated in our subsequent theoretical analysis, the generative error induced by this practical relaxation can be rigorously quantified, providing strong guarantees for the framework's effectiveness.

 Representation-based approaches to generative modeling have been widely studied, with prominent examples including variational autoencoders \citep{kingma2014auto},  vector-quantized generative models \citep{esser2021taming}, latent score-based  models \citep{vahdat2021score}, and latent diffusion models \citep{rombach2022high}.  Existing theoretical analyses of these frameworks typically adopt an autoencoder structure and characterize generative error through the reconstruction discrepancy of the encoder-decoder pair \citep{jiao2024latent, jiao2024convergence, chong2026semisupervise}. However, the theoretical guarantees for representation-based conditional generation remain largely unexplored. In particular, it remains unclear which structural properties a representation must possess to preserve the conditional distribution $P_{X \mid Y}$, and how information loss from an imperfect representation propagates into conditional generative error. A precise characterization of these relationships is essential. It establishes the fundamental limits of representation-based conditional generation and provides a principled basis for allocating limited labeled data across stages in the semi-supervised setting.

Our main contributions are summarized as follows:
\begin{enumerate}
	\item We propose RepG, a two-stage representation-based framework for \textit{semi-supervised conditional generation}. The structural factorization $Y \rightarrow R \rightarrow X$ separates the generative problem into a low-dimensional label-dependent stage, estimated from labeled data, and a high-dimensional label-free stage, estimated from the pooled sample.

	\item We establish a rigorous decomposition of the expected generative error of RepG into three components: the representation's structural bias, the error in learning the conditional distribution of the representation given the labels, and the error in generating data from the representation using all available data. This breakdown connects representation quality to the overall risk, providing a principled guide for choosing representations.

	\item  We establish convergence rates for the conditional generative error of RepG. Under smoothness assumptions, this error is bounded by a combination of mutual information, a vanishing time-truncation error, and a sample complexity term that depends on reduced dimensions rather than the full ambient dimension. We complement this upper bound with a theoretical lower bound showing that direct generative procedures inherently suffer from errors dependent on the full dimension. This theoretical gap confirms that our proposed factorization significantly improves sample complexity when the reduced dimension is much smaller than the ambient dimension.

	\item  We provide empirical validation  of RepG through simulation studies in both discrete and continuous conditioning scenarios, demonstrating that the proposed method effectively captures complex joint dependence structures. Applied to MNIST digit generation, RepG  yields  monotonic accuracy gains as the unlabeled sample size increases, validating our theoretical predictions.
\end{enumerate}

The remainder of the paper is organized as follows. Section~\ref{sec_model} introduces the representation-based conditional generative framework. Section~\ref{sec_ssl} presents the semi-supervised estimator RepG via conditional stochastic interpolation and details the  training procedure. Section~\ref{sec_pre}
contains the main theoretical results under semi-supervised and fully supervised settings. Section~\ref{sec_num} presents numerical studies. Finally, Section~\ref{sec_con} concludes with a discussion.  

  \paragraph{Notation.}
  Let $\mathbb{R}_{\ge 0}$ denote the non-negative real line and   $\mathbb{N}_0$   the set of non-negative integers.
  For $\bx\in\Rbb^d$,   $\|\bx\|_2$ denotes the Euclidean norm.
  Let $C(\mathbb{I}; \mathbb{R}^d)$ denote the space of continuous functions mapping from $\mathbb{I}$ to $\mathbb{R}^d$, equipped with the supremum norm.
  For random pair $(X, Y)$, $P_{X|Y}$ and $p_{X|Y}$ denote the conditional distribution and conditional density, respectively. For sequences $a_n$ and $b_n$, $a_n \lesssim b_n$ or $a_n = \mathcal{O}(b_n)$ denotes that $a_n \le C b_n$ for some constant $C > 0$.

  \section{Representation-based conditional generation}
  \label{sec_model}
 
 Consider a pair of random vectors $(X, Y) \in \mathbb{R}^d \times \mathbb{R}^q$, where the central goal is to characterize and sample from the conditional distribution $P_{X \mid Y}$. In this section, we introduce our proposed RepG approach for conditional generative learning, which relies on decomposing the conditional distribution via low-dimensional representations. Furthermore, we present an exact error decomposition in terms of the Kullback-Leibler (KL) divergence and elucidate how this error is governed by the sufficiency of the chosen representation.

 \subsection{Conditional decomposition via sufficient representations}

 When the ambient dimension $d$ is large, direct generation from $P_{X \mid Y}$ is statistically challenging, as the model must capture the complex dependence structure of a high-dimensional distribution from limited observations.

We address this difficulty by introducing a low-dimensional representation of the data. Let $R(\cdot): \mathbb{R}^d \to \mathcal{R} \subseteq \mathbb{R}^k$, with $k \ll d$, denote a fixed, deterministic representation mapping that is pre-specified independently of the observed data. The law of total probability yields the integral representation
$$
p_{X \mid Y}(\bx \mid \by) = \int_{\mathcal{R}} p_{X \mid R,Y}(\bx \mid \br, \by) \, p_{R \mid Y}(\br \mid \by) \, \mathrm{d}\br.
$$
This identity suggests a two-stage sampling procedure: first drawing $\br \sim P_{R \mid Y}(\cdot \mid \by)$, followed by $\bx \sim P_{X \mid R,Y}(\cdot \mid \br, \by)$. However, modeling the high-dimensional conditional distribution $P_{X \mid R, Y}$  requires labeled data  and does not  accommodate unlabeled data. In the semi-supervised setting considered in this work, it is crucial to leverage abundant unlabeled data during training.

To develop an efficient framework that naturally integrates unlabeled data, we require the representation $R$ to summarize all information in $X$ that is relevant to $Y$. We formalize this by imposing the conditional independence condition:
$
X \perp\!\!\!\perp Y \mid R.
$
Under this assumption, the conditional density simplifies to $p_{X \mid R,Y}(\bx \mid \br, \by) = p_{X \mid R}(\bx \mid \br)$, yielding the Markov chain factorization $Y \rightarrow R \rightarrow X$.
Based on this structural assumption, RepG partitions the conditional generation problem into two distinct components. The first component, the distribution $P_{R \mid Y}$, characterizes the dependence of the latent representation on the conditioning variable. Because $R$ takes values in $\mathbb{R}^k$ with $k \ll d$, this distribution can be estimated using limited labeled data. The second component, the distribution $P_{X \mid R}$, characterizes the high-dimensional data structure. Since $P_{X \mid R}$  is independent of $Y$, it can be learned using abundant unlabeled data.

Once a sufficient representation $R$ is available,
RepG defines the sampling process for $P_{X \mid Y}$ given any condition $Y = \by$ as follows:
\begin{enumerate}
    \item[(a)] \textit{Latent sampling:} Sample a representation $\br \sim P_{R \mid Y}(\cdot \mid \by)$.
    \item[(b)] \textit{Reconstruction:} Sample the target $\bx \sim P_{X\mid R}(\cdot \mid \br)$, bypassing the need to condition on $Y$ during the  generation phase.
\end{enumerate}

RepG is well-suited for \textit{semi-supervised learning}. By partitioning the generative process, we can leverage unlabeled observations of $X$ to learn the reconstruction distribution $P_{X \mid R}$. The scarce labeled pairs $(X, Y)$ are then used exclusively to estimating the low-dimensional latent distribution $P_{R \mid Y}$.  This structural separation significantly reduces the high sample complexity, circumventing the severe statistical burden inherent in modeling conditional distributions directly within the ambient space. Furthermore, when the conditional independence relation holds only approximately, this decomposition remains an effective structural approximation. Our subsequent theoretical analysis explicitly quantifies the resulting generative error induced by this approximation.

\subsection{Error decomposition}
To quantify the error induced by the conditional independence assumption, we analyze the discrepancy between the true conditional distribution and the model-implied conditional distributions. Let $P_{X,R \mid Y}$ denote the true joint conditional distribution, which factorizes exactly as $P_{X\mid R,Y} P_{R\mid Y}$. For a generic probability measure $P$, we let $\widehat{P}$ denote its estimated counterpart.

We compare the following two estimators of the joint conditional distribution.
The first  preserves the exact dependence structure:
$$
\widehat{P}^{\mathrm{Exact}}_{X,R \mid Y}
\coloneqq
\widehat{P}_{X \mid R,Y}\, \widehat{P}_{R \mid Y}.
$$
The second is the RepG that imposes the representation-based conditional factorization $Y \to R \to X$:
$$
\widehat{P}^{\mathrm{RepG}}_{X,R \mid Y}
\coloneqq
\widehat{P}_{X \mid R}\, \widehat{P}_{R \mid Y}.
$$
We measure the discrepancy between distributions using the KL divergence, defined for distributions $P$ and $Q$ with densities $p$ and $q$ as $\Dkl(P \| Q)=\int p(\bx) \log \{p(\bx)/q(\bx)\} \mathrm{d}\bx$. A central quantity in our analysis is the conditional mutual information (CMI):
\begin{equation*}
	I(X;Y \mid R) =	\mathbb{E}_{R}\!\left[D_{\mathrm{KL}}\left(	P_{X,Y \mid R}	\,\middle\|\,
	P_{X \mid R} \otimes P_{Y \mid R}\right)	\right],
\end{equation*}
which quantifies the residual dependence between $X$ and $Y$ given $R$. By definition, $I(X;Y \mid R)=0$ if and only if $X \inde Y \mid R$.

\begin{restatable}{theorem}{KLerror}
	\label{thm_KL_error_decom}
	The expected KL divergence between the true joint conditional distribution and the model-implied conditional distributions satisfies the following decomposition.
	\begin{enumerate}
\setlength\itemsep{-0.5em}
		\item[(a)] For the exact two-stage estimator,
		\begin{align*}
		&	\Ebb_{Y}\left[\Dkl(P_{X,R\mid Y} \,\|\, \widehat{P}^{\mathrm{Exact}}_{X,R\mid Y}) \right]\\
			&= \Ebb_{Y}\big[\Dkl(P_{R|Y} \,\|\, \widehat{P}_{R|Y})\big]
			+ \Ebb_{R,Y}\big[\Dkl(P_{X|R,Y} \,\|\, \widehat{P}_{X|R,Y}) \big].
		\end{align*}
		\item[(b)] For the proposed representation-based estimator RepG,
		\begin{align*}
			&\Ebb_{Y}\left[\Dkl(P_{X,R\mid Y} \,\|\, \widehat{P}^{\mathrm{RepG}}_{X,R\mid Y}) \right]\\
			&= I(X;Y|R) + \Ebb_{Y}\big[D_{KL}(P_{R|Y} \,\|\, \widehat{P}_{R|Y})\big]
			+ \Ebb_{R} \big[ D_{KL}(P_{X|R} \,\|\, \widehat{P}_{X|R})\big].
		\end{align*}
	\end{enumerate}
\end{restatable}

Theorem \ref{thm_KL_error_decom} provides a precise decomposition of the total error of RepG into a structural approximation component and a statistical estimation term. In part (a), the error consists entirely of the estimation errors for $P_{R\mid Y}$ and $P_{X\mid R,Y}$. In part (b), the decomposition for the representation-based model introduces the CMI term $I(X;Y\mid R)$. This  characterizes the structural approximation error, quantifying the loss of predictive information about $Y$ when conditioning only on the representation $R$. The remaining two terms correspond to the estimation error of the low-dimensional and high-dimensional components, respectively. Consequently, strictly satisfying the conditional independence condition is not necessary for practical utility. Provided that $I(X;Y\mid R)$ is small and the corresponding estimation errors are appropriately controlled, the total distribution error remains small.

\subsection{Connection to sufficient dimension reduction}
\label{sec_sdr_connection}

Theorem~\ref{thm_KL_error_decom} identifies $I(X;Y \mid R)$ as the structural approximation error that is induced by the representation constraint. The effectiveness of the proposed generative framework therefore depends on constructing a representation mapping $R(\cdot)$ that makes this quantity small.
This objective connects our framework to the literature on sufficient dimension reduction  \citep[SDR;][]{Li1991, cook2000SAVE, Lee2013, HuangSRL2024, Deepnonlinear}.

Since $R(X)$ is a deterministic function of $X$, by the chain rule for mutual information, we obtain the identity
\[
I(X;Y) = I(R(X);Y) + I(X;Y \mid R(X)).
\]
The mutual information $I(X;Y)$ depends only on the underlying data-generating process and remains invariant to the choice of representation. Consequently, maximizing the captured predictive information $I(R(X);Y)$ is equivalent to minimizing the residual dependence $I(X;Y \mid R(X))$.  A representation is sufficient for predicting $Y$ if and only if $I(X;Y \mid R(X))=0$,  implying that $R(X)$  preserves all predictive information about $Y$.

This relationship justifies adopting existing SDR criteria within our generative framework. Although classical SDR procedures focus on linear representations and typically optimize surrogate measures of dependence, their population-level objectives are closely related to learning representations that reduce this residual dependence. For instance, \citet{HuangSRL2024} propose learning $R$ by maximizing the distance covariance   $\mathcal{V}(R(X), Y)$ over a constrained class
 $\mathcal{M} = \{R: \mathbb{R}^d \to \mathbb{R}^k \mid R(X) \sim \mathcal{N}(0, \bI_k)\}$. Their analysis demonstrates that, under suitable conditions, global maximizers of this objective recover sufficient representations. This property makes their criterion a computationally feasible surrogate for minimizing $I(X; Y \mid R(X))$.  We employ this approach for constructing $R$ in the numerical studies of Section~\ref{sec_num}.

\section{Semi-supervised conditional generation}
\label{sec_ssl}
We now consider the semi-supervised setting, where labeled paired observations $\{(X_i, Y_i)\}$ are scarce, yet unlabeled target samples $\{\widetilde{X}_j\}$ are abundantly available. This regime arises frequently in practice, as obtaining labels $Y$ often requires costly expert annotation or delayed measurements, whereas the large-scale collection of unlabeled observations $X$ is relatively easy.

Direct conditional generative modeling relies entirely on the labeled sample $\mathcal{D}_l$ and, therefore, cannot readily benefit from the unlabeled data $\mathcal{D}_u$. The representation-based factorization of RepG introduced in Section~\ref{sec_model} resolves this limitation by decoupling the generative process into a label-dependent, low-dimensional component and a label-free, high-dimensional component. This separation enables the effective use of both data sources during training, drastically reducing the statistical burden of learning high-dimensional conditional structures.

Specifically, suppose we observe
$
\mathcal{D}_l = \{(X_i, Y_i)\}_{i=1}^n$ and $ \mathcal{D}_u = \{\widetilde{X}_j\}_{j=1}^N,
$
where $N \gg n$, and the unlabeled observations are drawn from the same marginal distribution of $X$ as in $\mathcal{D}_l$. The generative procedure of RepG is carried out in two consecutive stages as  below.
\begin{itemize}
	\item[]  {Stage 1 ($P_{R|Y}$).} This stage characterizes the conditional distribution governing the latent representation $R \in \mathcal{R} \subseteq \mathbb{R}^k$. Because this component is conditioned on $Y$, it must be estimated using the labeled pairs $\{(R_i,Y_i)\}_{i=1}^n$, with $R_i=R(X_i)$. The low intrinsic dimensionality $k \ll d$ ensures stable estimation even when the labeled sample size $n$ is limited.
	\item[]    {Stage 2 ($P_{X|R}$):} The second stage focuses on the conditional distribution mapping the latent representation back to the target space $\mathbb{R}^d$. By the conditional independence assumption, this component does not depend on $Y$. It can therefore be estimated from the pooled augmented dataset $\mathcal{D}_{\mathrm{all}} = \{(X_i,R_i)\}_{i=1}^n \cup \{(\widetilde X_j,\widetilde {R}_j)\}_{j=1}^N$ of size $n+N$, with $\widetilde{R}_j=R(\widetilde X_j)$.
\end{itemize}

This factorization of RepG can yield a substantial statistical advantage. It partitions a challenging high-dimensional supervised problem into two tractable estimation tasks, systematically  allocating the scarce labeled observations to the low-dimensional representation learning and leveraging the abundant unlabeled data for the high-dimensional target modeling.

\subsection{Implementation via conditional stochastic interpolation}

To implement the proposed two-stage generation RepG framework, we adopt conditional stochastic interpolation \citep{albergo2023stochastic, huang2023conditional}. This construction is well-suited  for our setting, as it converts conditional generative modeling into a sequence of least-squares nonparametric regression problems, thereby permitting both scalable computation and direct statistical analysis.

For a generic target variable $Z_1 \sim P$ and Gaussian reference variable $Z_0\sim\mathcal{N}( {0},\bI)$, define the interpolation path
 \begin{equation} \label{eq_interp}
 	Z_t=a_t Z_0+b_t Z_1,\qquad t\in[0,1],
 \end{equation}
 where the deterministic schedules satisfy $a_0=b_1=1$ and $a_1=b_0=0$. The path therefore evolves continuously from the tractable reference distribution at $t=0$ to the target distribution at $t=1$. Under mild regularity conditions, the marginal density of $Z_t$, denoted $p_t$, can be generated by the stochastic  differential equation (SDE)
 \begin{equation}\label{eq_genSDE}
 	\diff Z_t=\bv_F^*(t,Z_t)\diff t+\sqrt{2\eta(t)}\,\diff B_t,
 	\qquad Z_0\sim\mathcal{N}( {0},\bI),
 \end{equation}
 where $\eta(t)\ge0$ is a pre-specified diffusion level. Simulating this SDE to $t=1$ yields a sample from the
 target distribution $P$, so the entire generative task reduces to estimating the  drift $\bv_F^*$. The   drift admits the decomposition
 \[
 \bv_F^*(t,\bz)=\bv^*(t,\bz)+\eta(t)\bs^*(t,\bz),
 \]
 with $\bv^*$ denoting the velocity field and $\bs^*(t,\bz)=\nabla_{\bz}\log p_t(\bz)$ the score function.

 A key advantage of this formulation is that the velocity field is characterized as the unique minimizer of the quadratic risk
 \begin{equation}\label{eq_loss_general}
 	\mathcal L(\bv)=
 	\mathbb E \left[	\big\|\dot a_T Z_0+\dot b_T Z_1-\bv(T,Z_T)
 	\big\|_2^2\right], 	\qquad T\sim \mathrm U(0,1).
 \end{equation}
 Hence, learning the generator reduces to least-squares regression based on paired samples $(Z_0,Z_1)$, without
 requiring evaluation of the generally intractable path density $p_t$. Moreover, given an estimator $\widehat{\bv}$, the corresponding score estimator can be recovered algebraically. This yields the plug-in drift
 \begin{equation}	\label{eq_vF}
 	\widehat{\bv}_F(t,\bz)
 	=	\widehat{\bv}(t,\bz)
 	+	\frac{\eta(t)}{a_t^2\partial_t\{\log(b_t/a_t)\}}
 	\left(	\widehat{\bv}(t,\bz)-\frac{\dot b_t}{b_t}\bz	\right).
 \end{equation}
 This explicit drift estimator depends only on the velocity field $\widehat{\bv}$ and the pre-specified interpolation schedules $a_t$ and $b_t$. The construction of the plug-in drift \eqref{eq_vF} follows from the Gaussian structure of the conditional interpolant, a relationship closely connected to Tweedie's formula and discussed in \citet{ gao2024gaussian,albergo2023stochastic}.

 We apply this conditional stochastic interpolation construction separately to the two components of the proposed factorization, where Stage 1 targets the conditional latent distribution $P_{R|Y}$ and Stage 2 targets the reconstruction distribution $P_{X|R}$.

\subsection{Velocity field estimation}
\label{sec_fnn_def}

To minimize the empirical counterpart of the risk \eqref{eq_loss_general}, we parameterize the velocity fields using deep ReLU feedforward neural networks (FNNs).

 A scalar-valued FNN $g: \mathbb{R}^{d_0} \to \mathbb{R}$ of depth $L$ can be expressed as a composition of functions  $g = g_{L} \circ g_{L-1} \circ \cdots \circ g_{0}$,  where $g_{i}(\bx)=\sigma(A_i\bx+\bb_i)$ for $i=0,\dots, L-1$, and the final linear layer is $g_{L}(\bx)=A_{L}\bx+b_{L}$. Here,  $A_{i}\in\mathbb{R}^{d_{i+1}\times d_{i}}$ and $\bb_{i}\in\mathbb{R}^{d_{i+1}}$ denote the weight matrices and bias vectors, and $\sigma(x)=\max\{x,0\}$ is the ReLU activation function applied component-wise. Let $\mathcal{NN}(W, L)$ denote the class of such scalar FNNs with width $W=\max\{d_1,\dots, d_L\}$ and depth $L$. To ensure uniform boundedness for  stability, we apply a truncation operator $T_{M_n}(y) = \max\{-M_n, \min(y, M_n)\}$. The vector-valued estimation class, with output dimension $s = k$ in Stage 1 and $s = d$ in Stage 2, is defined by
\begin{equation*}
	\cF_n(W,L,s) = \Big\{\bbf=(f_1,\dots, f_s)^{\top}:\;  f_j\in T_{M_n}\{\mathcal{NN}(W,L)\}  \text{ for } j\in[s]\ \Big\}.
\end{equation*}

We now formulate the estimation objectives for the two generative stages. In the first stage, the latent velocity field $\bu^*:[0,1]\times\mathbb{R}^k\times\mathbb{R}^q\to\mathbb{R}^k$ is estimated using   the labeled pairs $\{(R_i, Y_i)\}_{i=1}^n$. We obtain the estimator $\widehat{\bu}_n \in \cF_n(W_1,L_1, k)$ by minimizing the empirical risk
\begin{equation}\label{eq_loss1}
	\mathcal{L}_n^{(1)}(\bu) = \frac{1}{n}\sum_{i=1}^n
	\big\|\dot{a}_{t_i}W_{0,i} + \dot{b}_{t_i}R_i
	- \bu(t_i, W_{t_i}; Y_i)\big\|_2^2,
\end{equation}
where $t_i \sim \mathrm{U}(0,1)$, $W_{0,i} \sim \mathcal{N}(0, \bI_k)$ are sampled independently for each
$i \in [n]$, and $W_{t_i} = a_{t_i}W_{0,i} + b_{t_i}R_i$ follows from the interpolation \eqref{eq_interp}.

In the second stage, the high-dimensional velocity field  $\bv^*:[0,1]\times\mathbb{R}^d\times\mathbb{R}^k\to\mathbb{R}^d$ is estimated from the augmented dataset $\mathcal{D}_{\mathrm{all}}$. Let $\{(X_m, R_m)\}_{m=1}^{n+N}$ denote the elements of $\mathcal{D}_{\mathrm{all}}$. We obtain the estimator $\widehat{\bv}_{n+N} \in \cF_n(W_2,L_2, d)$ by minimizing the empirical risk
\begin{equation}\label{eq_loss2}
	\mathcal{L}_{n+N}^{(2)}(\bv) = \frac{1}{n+N}
	\sum_{m=1}^{n+N}
	\big\|\dot{a}_{t_m}Z_{0,m} + \dot{b}_{t_m}X_m
	- \bv(t_m,Z_{t_m};R_m)\big\|_2^2,
\end{equation}
where $t_m \sim \mathrm{U}(0,1)$, $Z_{0,m} \sim \mathcal{N}(0, \bI_d)$ are sampled independently for each $m \in [n+N]$, and $Z_{t_m} = a_{t_m}Z_{0,m} + b_{t_m}X_m$ follows from \eqref{eq_interp}.

\begin{algorithm}[t]
	\caption{Semi-Supervised Representation-Based
		Conditional Generation}
	\label{algo_full}
	\begin{algorithmic}[1]
		\Require Labeled data $\mathcal{D}_l = \{(X_i, Y_i)\}_{i=1}^n$, unlabeled data $\mathcal{D}_u = \{\widetilde{X}_j\}_{j=1}^N$, representation map $R(\cdot)$, time grid $\{\tau_h\}_{h=0}^H$, condition $\by$.
		\Ensure Generated sample $\bar{\bx}$ approximately distributed as $P_{X \mid Y=\by}$.
		
		\vspace{1.5mm}
		\Statex \textit{Phase I: Semi-supervised training}
		\State Compute latent features $\{R_i\}_{i=1}^n$ and $\{\widetilde{R}_j\}_{j=1}^N$ via $R(\cdot)$.
		\State Obtain $\widehat{\bu}_n$ by minimizing $\mathcal{L}_n^{(1)}$ over $\{(R_i, Y_i)\}_{i=1}^n$.
		\State Obtain $\widehat{\bv}_{n+N}$ by minimizing $\mathcal{L}_{n+N}^{(2)}$ over $\mathcal{D}_{\mathrm{all}} = \{(X_i,R_i)\} _{i=1}^n\cup \{(\widetilde{X}_j,\widetilde{R}_j)\}_{j=1}^N$.
		\State Construct forward drift fields $\widehat{\bu}_{n,F}$ and $\widehat{\bv}_{n+N,F}$ via \eqref{eq_vF}.
		
		\vspace{1.5mm}
		\Statex \textit{Phase II: Two-stage sequential generation}
		\State Draw $\bar{W}_0 \sim \mathcal{N}(0, \bI_k)$.
		\For{$h = 1, \dots, H$}
		\State Sample $\Delta B_h \sim \mathcal{N}\left(0, 	\int_{\tau_{h-1}}^{\tau_h} 2\eta(s)\diff s \cdot \bI_k\right)$.
		\State $\bar{W}_{\tau_h} \gets \bar{W}_{\tau_{h-1}} + \widehat{\bu}_{n,F}(\tau_{h-1}, \bar{W}_{\tau_{h-1}};\by)(\tau_h - \tau_{h-1}) + \Delta B_h$.
		\EndFor
		\State Set $\hat{\br} = \bar{W}_{1-\varepsilon}$.
		
		\State Draw $\bar{Z}_0 \sim \mathcal{N}(0, \bI_d)$.
		\For{$h = 1, \dots, H$}
		\State Sample $\Delta \widetilde{B}_h \sim \mathcal{N}\left(0, \int_{\tau_{h-1}}^{\tau_h} 2\eta(s)\diff s \cdot \bI_d\right)$.
		\State $\bar{Z}_{\tau_h} \gets \bar{Z}_{\tau_{h-1}} + \widehat{\bv}_{n+N,F}(\tau_{h-1}, \bar{Z}_{\tau_{h-1}};\hat{\br})(\tau_h - \tau_{h-1}) + \Delta \widetilde{B}_h$.
		\EndFor
		\State \textbf{Return} $\hat{\bx} = \bar{Z}_{1-\varepsilon}$.
	\end{algorithmic}
\end{algorithm}

\subsection{Sequential generation}

Given the network estimators $\widehat{\bu}_n$ and $\widehat{\bv}_{n+N}$, the corresponding drift fields $\widehat{\bu}_{n,F}$ and $\widehat{\bv}_{n+N,F}$ are constructed via \eqref{eq_vF}. For a given condition $Y=\by$, the latent representation is first generated by simulating
\begin{equation}\label{eq_EM1}
	\diff \bar{W}_t = \widehat{\bu}_{n,F}(t, \bar{W}_t; \by)\diff t + \sqrt{2\eta(t)}\diff B_t, \qquad 	\bar{W}_0 \sim \mathcal{N}(0, \bI_k),
\end{equation}
where $\{B_t\}_{t\geq 0}$ is a standard Brownian motion in $\mathbb{R}^k$. The latent representation is extracted at the truncated time $1-\varepsilon$ as $\hat{\br} = \bar{W}_{1-\varepsilon}$ for a small $\varepsilon\in(0,1)$, avoiding the terminal singularity of the drift field. Conditioned on $\hat{\br}$, the high-dimensional target is subsequently generated by simulating
\begin{equation}\label{eq_EM2}
	\diff \bar{Z}_t = \widehat{\bv}_{n+N,F}(t, \bar{Z}_t; \hat{\br})\diff t + \sqrt{2\eta(t)}\diff \widetilde{B}_t, \qquad \bar{Z}_0 \sim \mathcal{N}(0, \bI_d),
\end{equation}
where $\{\widetilde{B}_t\}_{t\geq 0}$ is a standard Brownian motion in $\mathbb{R}^d$, independent of $\{B_t\}_{t\geq 0}$, and the final generated sample is taken as $\hat{\bx} = \bar{Z}_{1-\varepsilon}$. In practice, both SDEs are solved via a discretization scheme over the time grid $0 = \tau_0 < \tau_1 < \cdots < \tau_H = 1-\varepsilon$, as detailed in Algorithm~\ref{algo_full}.

\section{Theoretical results}

We now establish theoretical guarantees for the semi-supervised generative procedure RepG described in Section~\ref{sec_ssl}. First, we derive error bounds for the nonparametric estimation of the velocity fields, and subsequently establish bounds for the semi-supervised conditional generation.  We also demonstrate that incorporating unlabeled data improves the convergence of conditional generative learning.

\subsection{Error analysis in velocity estimation}
  \label{sec_pre}

This subsection establishes error bounds for the estimated velocity functions, laying the groundwork for the error analysis in conditional generation.

\subsubsection{Generative path measures}\label{sec_constr}

To ensure a smooth transition from the prior distribution to the target and stable regression targets, the schedule functions $a_t$ and $b_t$ in \eqref{eq_interp} are required to satisfy the following regularity conditions.
\begin{assumption} \label{ass_coeffs}
 	The functions $a_t, b_t : [0,1] \to \mathbb{R}$ are continuously differentiable and satisfy the boundary conditions $a_0 = 1,  b_0 = 0,$ and $ a_1 = 0,   b_1 = 1.$ Moreover, for all $t \in (0,1)$, $	\dot a_t < 0,   \dot b_t > 0, $ and  $\sup_{t \in [0,1]}\max\{|\dot{a}_t|, |\dot b_t|\} \leq C$ for some constant $C>0$.
\end{assumption}
A standard example satisfying these conditions is the linear schedule $a_t = 1-t$ and $b_t = t$, which is used in rectified flow models \citep{liu2023flow}.

We now proceed to bound the KL divergence between the true and estimated generative distributions. Evaluating this error directly at the terminal time $t=1$ poses significant analytical challenges. The marginal distribution lacks a tractable closed form, and more critically, the weight function $w(t) = \{a_t^2\,\partial_t\log(b_t/a_t)\}^{-1}$ in \eqref{eq_vF} diverges as $t \to 1$ due to the boundary condition $a_1=0$ and the boundedness of the derivatives. To overcome this singularity, we lift our analysis to the truncated path space $\Omega = C([0,1-\varepsilon]; \mathbb{R}^d)$ for a small $\varepsilon \in (0,1)$. The following lemma formalizes the path-space error bound for the Stage 2 SDE.

\begin{restatable}{lemma}{KLboundstage}
	\label{thm_KL_bound_stage2}
	Let $\bv^*:[0,1]\times\mathbb{R}^d\times\mathbb{R}^k\to\mathbb{R}^d$ be the true velocity field and $\widehat{\bv}$ a continuous uniformly bounded approximation. Define the corresponding  drifts $\bv_F^*$ and $\widehat{\bv}_F$ via the
	plug-in relation \eqref{eq_vF}. Consider the true and estimated generative processes $\{Z_t\}$ and $\{\bar{Z}_t\}$, governed by the SDEs
	\begin{align*}
		\diff Z_t &= \bv_F^*(t, Z_t; R)\diff t + \sqrt{2\eta(t)}\diff B_t, \qquad
		Z_0\sim\mathcal{N}(0,\bI_d),\\	
		\diff \bar{Z}_t &= \widehat{\bv}_F(t,\bar{Z}_t; R)\diff t + \sqrt{2\eta(t)}\diff B_t, \qquad \bar{Z}_0\sim\mathcal{N}(0,\bI_d),
	\end{align*}
	where $\{B_t\}_{t\geq 0}$ is a standard Brownian motion in $\mathbb{R}^d$ and $R\in\mathbb{R}^k$ is a fixed conditioning variable. Let $\mathbf{P}_{|R}$ and $\widehat{\mathbf{P}}_{|R}$ denote the conditional path measures of $\{Z_t\}$ and $\{\bar{Z}_t\}$ on $\Omega = C([0,1-\varepsilon];\mathbb{R}^d)$, and let $\gamma(t) = \{4\eta(t)\}^{-1}(1+\eta(t)w(t))^2$ where $w(t)=\{a_t^2\partial_t\log(b_t/a_t)\}^{-1}$. Assume $\bv^*$ satisfies the exponential integrability condition
	\begin{equation}\label{eq_exp_integrability}
		\Ebb_{\mathbf{P}_{|R}}\left[\exp\left(\int_0^{1-\varepsilon} 2\gamma(t)\big\|\bv^*(t,Z_t;R)\big\|_2^2\diff t\right)\right] < \infty.
	\end{equation}
	Then the expected path-space KL divergence satisfies
	\begin{equation*}
		\Ebb_{R}\big[\Dkl(\mathbf{P}_{|R}\parallel \widehat{\mathbf{P}}_{|R})\big] = (1-\varepsilon)\Ebb_{T,Z_T,R}\left[
		\frac{1}{4\eta(T)}\big\|\widehat{\bv}_F(T,Z_T;R) - \bv_F^*(T,Z_T;R)\big\|_2^2\right],
	\end{equation*}
	where $T\sim\mathrm{U}(0,1-\varepsilon)$. Furthermore, the expected marginal KL divergence at truncation time $1-\varepsilon$ satisfies
	\begin{equation*}
		\Ebb_{R}\big[\Dkl(P_{Z_{1-\varepsilon}|R}\parallel  P_{\bar{Z}_{1-\varepsilon}|R})\big] \leq (1-\varepsilon)\Ebb_{T,Z_T,R}	\left[\gamma(T)\big\|\widehat{\bv}(T,Z_T;R) - \bv^*(T,Z_T;R)\big\|_2^2\right].
	\end{equation*}
\end{restatable}

Lemma~\ref{thm_KL_bound_stage2} applies Girsanov's theorem~\citep{Gall2016} to convert the path-space KL divergence into an integrated $L_2$-estimation error of the drift fields. The exponential integrability condition in \eqref{eq_exp_integrability} is a standard regularity assumption for the change-of-measure argument. It ensures absolute continuity between the corresponding path measures, under which the KL divergence admits the stated quadratic representation.

Beyond this continuous-time modeling error, numerical time discretization introduces an additional source of discrepancy.

  \begin{restatable}{corollary}{propnumerror}
  	\label{prop_num_error}
  	Let $0 = \tau_0 < \tau_1 < \dots < \tau_H = 1-\varepsilon$ be a   time grid with truncation mapping $\phi(t) = \tau_{h-1}$ for $t \in [\tau_{h-1}, \tau_h)$. Under the conditions of Lemma \ref{thm_KL_bound_stage2}, let $\bar{\mathbf{P}}_{|R}$  denote the path measure on $\Omega$ induced by the discretization $\diff \bar{Z}_t = \widehat{\bv}_F(\phi(t), \bar{Z}_{\phi(t)}; R)\diff t + \sqrt{2\eta(t)}\diff B_t$.   Then, the expected path-space   divergence between the true and discretized dynamics is bounded by
  	\begin{align*}
  		\Ebb_{R}\big[\Dkl(\mathbf{P}_{|R} \parallel \bar{\mathbf{P}}_{|R})\big] \le &\int_0^{1-\varepsilon} \Ebb_{Z_t, R}\left[ \frac{1}{2\eta(t)} \big\| \widehat{\bv}_{F}(t, Z_t; R) -\bv_F^*(t, Z_t; R) \big\|_2^2 \right] \diff t \\
  		&+ \sum_{h=1}^H \int_{\tau_{h-1}}^{\tau_h} \Ebb_{Z_t, R}\left[ \frac{1}{2\eta(t)} \big\|\widehat{\bv}_{F}(t, Z_t; R) - \widehat{\bv}_{F}(\tau_{h-1}, Z_{\tau_{h-1}}; R) \big\|_2^2 \right] \diff t.
  	\end{align*}
  \end{restatable}

The second term in the bound of Corollary \ref{prop_num_error} captures the time discretization error, which becomes negligible for a sufficiently fine time grid under standard regularity conditions \citep{Lu2023,chen2023sampling}.    Consequently, provided the grid is sufficiently fine, the discretization error becomes asymptotically negligible. The total distributional discrepancy is thus dominated by the velocity estimation error, which motivates our primary focus on the nonparametric estimation of the velocity fields.

\subsubsection{Error bounds for estimated velocity}\label{sec_velocityesti}

We now establish convergence rates for the velocity estimators, which requires imposing precise smoothness conditions on the target fields.
 \begin{definition}[H{\"o}lder classes] \label{def_holder}
 	Let $p\in\mathbb{N}$ and $\tau>0$ with $\tau = \underline{\tau} + \rho$ for $\underline{\tau} \in \mathbb{N}_0$ and $\rho \in (0, 1]$. For a subset $ \mathbb{I} \subseteq \mathbb{R}^p$ and a constant $B> 0$, the H\"{o}lder class $\mathcal{H}^{\tau}(\mathbb{I}, B)$ with smoothness level $\tau$ is defined as
 	\begin{equation*}
 		\mathcal{H}^{\tau}(\mathbb{I}, B)=	\bigg\{	f: \mathbb{I} \rightarrow \Rbb \big| \max_{\|\bs\|_{1} \leq \underline{\tau}} \sup_{\bx \in \mathbb{I}} |\partial^{\bs} f(\bx)| \leq B,   \; \max _{\|\bs\|_{1}= \underline{\tau}} \sup _{\bx_1 \neq \bx_2 \in \mathbb{I}} \frac{\left|\partial^{\bs} f(\bx_1)-\partial^{\bs} f(\bx_2)\right|}{\|\bx_1-\bx_2\|_{2}^{\rho}} \leq B\bigg\},
 	\end{equation*}
 	where $\partial^{\bs}=\partial^{s_1}\cdots\partial^{s_p}$ for $\bs=(s_1,\dots,s_p)^\top\in\mathbb{N}_0^p$.
 \end{definition}

In diffusion and flow-based generative models, assuming global H\"older smoothness is typically overly restrictive, because the velocity fields exhibit unbounded behavior under standard conditions. For full-support distributions, velocity fields typically lack a uniform global bound, as exemplified by the linear growth of Gaussian velocity fields \citep{hertrich2025on}. Conversely,   distributions with restricted support induce terminal singularities, forcing the velocity field to diverge to infinity as $t \to 1$ for points outside the data support \citep{wan2025elucidating}. To accommodate these structural properties, we relax the global constraint and impose a localized smoothness condition. Specifically, we assume the velocity fields belong to a H\"older class over compact hypercubes, allowing the local H\"older norm to grow at most polynomially with the domain radius. This relaxed condition is   satisfied by Gaussian velocity fields.

  \begin{assumption}\label{ass_smoothness}
  	There exist constants $\beta,\alpha > 0$ and a polynomial function $P(\cdot)$ such that for any radius $ M> 0$ and hypercube $\mathbb{I}_M = [-M, M]^p$, the target velocity components satisfy $u_j^* \in \mathcal{H}^\beta(\mathbb{I}_M, P(M))$ for $j \in [k]$ with $p = 1+k+q$, and $v_j^* \in \mathcal{H}^\alpha(\mathbb{I}_M, P(M))$ for $j \in [d]$ with $p = 1+k+d$.
  \end{assumption}

  Evaluating the global estimation error under local smoothness conditions requires controlling tail contributions from data outside compact domains. To this end, we assume the underlying distributions satisfy the following subexponential moment bounds.

  \begin{assumption} \label{ass_tail}
  	There exist constants $\sigma_{X},\sigma_{R},\sigma_{Y}> 0$ such that $\max_{j \in [d]} \mathbb{E} \big[ \exp \big\{\sigma_{X} |X^{(j)}| \big\} \big] < \infty$, $\max_{j\in[k]}\Ebb \big[\exp \big\{\sigma_{R} |R^{(j)}|\big\} \big] < \infty$, and $\max_{j\in[q]}\Ebb \big[\exp \big\{\sigma_{Y} |Y^{(j)}|\big\} \big] < \infty$.
  \end{assumption}

  Under Assumption~\ref{ass_tail}, restricting the target domain to a compact hypercube with a radius proportional to the logarithm of the stage-specific sample size ensures that the approximation error in the tail regions is statistically negligible. Combining this truncation strategy with the approximation capacity of FNNs yields the following non-asymptotic convergence rates.

  \begin{restatable}{theorem}{thmFNN}\label{Thm_FNN}
  	Suppose Assumptions \ref{ass_coeffs}--\ref{ass_tail} hold. Let the network architecture for the first-stage estimator $\widehat{\bu}_n$ be constructed such that the width and depth product scales as $W_1L_1 \asymp n^{\frac{1+q+k}{2(1+q+k)+4\beta}}\log n$. Then,
  	\[
  	\mathbb{E}  \Big[\big\| \widehat{\bu}_n(T,a_TW_0+b_TR;Y) - \bu^{*}(T,a_TW_0+b_TR;Y)  \big\|_2^2\Big] \lesssim n^{-\frac{2\beta}{2\beta + q + k + 1}} \mathrm{poly}(\log n),
  	\]
  	where   $\mathrm{poly}(\log n)$ denotes a polylogarithmic factor bounded by $c (\log n)^r$ for some universal constants $c, r > 0$ independent of the sample size $n$. Furthermore, let $W_2L_2 \asymp (n+N)^{\frac{1+d+k}{2(1+d+k)+4\alpha}}\log (n+N)$ for the second-stage estimator $\widehat{\bv}_{n+N}$. Then,
  	\[
  	\mathbb{E}\Big[\big\| \widehat{\bv}_{n+N}(T,a_TZ_0+b_TX;R)  - \bv^{*} (T,a_TZ_0+b_TX;R) \big\|_2^2\Big] \lesssim (n+N)^{-\frac{2\alpha}{2\alpha + d + k + 1}} \mathrm{poly}(\log (n+N)).
  	\]
  \end{restatable}
  Theorem~\ref{Thm_FNN} establishes that the FNN estimators achieve the minimax optimal rate for nonparametric regression \citep{Stone1982}, up to polylogarithmic factors. Unlike standard theoretical results for deep ReLU networks \citep[see, e.g.,][]{Schmidt2020,Kohler2021,jiao2023deep} that assume globally bounded target functions, our framework explicitly accommodates the localized smoothness and tail-bound conditions required to handle the inherent unboundedness of velocity fields.

\subsection{Error bounds for conditional generation}
\label{sec_main}

This subsection establishes end-to-end statistical guarantees for the proposed framework. We derive convergence rates for both the semi-supervised and fully supervised settings, and establish a minimax lower bound for direct conditional generation. Together, these results show that the representation-based framework achieves statistical advantages over direct conditional sampling.

\subsubsection{Semi-supervised setting}
\label{sec_semi}

We first establish the convergence rate of the proposed semi-supervised generative estimator. Recall that the generative procedure summarized in Algorithm~\ref{algo_full} proceeds sequentially. The first-stage SDE~\eqref{eq_EM1} generates a latent representation targeting $P_{R \mid Y}$ using the velocity field estimated from the labeled data $\mathcal{D}_l$, and the second-stage SDE~\eqref{eq_EM2} reconstructs the observation targeting $P_{X \mid R}$ using the augmented dataset $\mathcal{D}_{\mathrm{all}}$ of size $n+N$.

To circumvent the analytical singularity at the terminal time $t=1$, where the weight function $\gamma(t)$ diverges, we follow standard practice in diffusion-based generative models \citep{chen2023sampling, jain2026a} and evaluate the distributional discrepancy at a truncated time $1-\varepsilon$ for a sufficiently small $\varepsilon \in (0,1)$. We define the early-stopped target variables as
$
	\widetilde{X} = a_{1-\varepsilon} Z_0 + b_{1-\varepsilon} X,
	\widetilde{R} = a_{1-\varepsilon} W_0 + b_{1-\varepsilon} R,
$
where $Z_0 \sim \mathcal{N}(0, \bI_d)$ and $W_0 \sim \mathcal{N}(0, \bI_k)$ are independent standard Gaussian vectors.   As $\varepsilon \to 0$, both $\widetilde{X}$ and $\widetilde{R}$ converge in distribution to $X$ and $R$, respectively.

 Let $\widetilde{R}' = R + \sigma_{\varepsilon} W_0,$ where
 $\sigma_{\varepsilon} = a_{1-\varepsilon}/b_{1-\varepsilon}$, and $W_0 \sim \mathcal{N}(\boldsymbol{0}, \bI_k)$ is independent of $(X, R, Y)$.
 Denote
$ \delta(\varepsilon) \coloneqq  I(R; Y) - I(\widetilde{R}'; Y).$
The following lemma shows that the structural approximation error of the early-stopped variables remains controlled by  $I(X;Y\mid R)$.

\begin{restatable}{lemma}{lemCMI}
	\label{lem_cmi}
	Suppose Assumption \ref{ass_coeffs} holds and $P_{X,R,Y}$ has finite second moments. Then,
$
		I(\widetilde{X};\,Y \mid \widetilde{R})
		\leq I(X;\,Y \mid R) +\delta(\varepsilon),
$
and  $\delta(\varepsilon)\downarrow0$ as $\varepsilon\downarrow0.$
\end{restatable}

Combining the KL decomposition in Theorem \ref{thm_KL_error_decom} with the estimation rates established in Section~\ref{sec_pre}, we obtain the following upper bound on the total generative error. Here, $\widehat{P}^{\,\mathrm{semi}}_{X\mid Y}$ denotes the conditional distribution of the final output induced by the sequential generative procedure detailed in Section \ref{sec_ssl}.
\begin{restatable}{theorem}{thmsemi}
	\label{thm_ssl}
	Suppose Assumptions \ref{ass_coeffs}--\ref{ass_tail} hold, and assume  $\bu^*$ and $\bv^*$ satisfy the exponential integrability conditions. Let $\varepsilon \in (0,1)$. Then, the expected KL divergence between the early-stopped target distribution and the semi-supervised generative distribution satisfies
	\begin{equation*}
		\mathbb{E}_{\mathcal{D}_l,\, \mathcal{D}_u,\, Y}\bigl[ D_{\mathrm{KL}}\bigl( P_{\widetilde{X} \mid Y} \,\|\, \widehat{P}^{\,\mathrm{semi}}_{X \mid Y} \bigr) \bigr]
		\leq I(\widetilde{X};\, Y \mid \widetilde{R}) + C_{\varepsilon} \Bigl\{ \mathcal{E}_{\bu}(n) + \mathcal{E}_{\bv}(n+N) \Bigr\},
	\end{equation*}
where $C_{\varepsilon} = \mathcal{O}\bigl(\sup_{t \in [0,\,1-\varepsilon]} \gamma(t)\bigr)$, and  $\mathcal{E}_{\bu}(n)$ and $\mathcal{E}_{\bv}(n+N)$ denote the  $L_2$-estimation errors of the first-stage and second-stage velocity fields, respectively. Under the network architectures specified in Theorem \ref{Thm_FNN}, these errors  satisfy
	\begin{equation*}
		\mathcal{E}_{\bu}(n) = n^{-\frac{2\beta}{2\beta + q + k + 1}} \operatorname{poly}(\log n), \quad \mathcal{E}_{\bv}(n+N) = (n+N)^{-\frac{2\alpha}{2\alpha + d + k + 1}} \operatorname{poly}(\log(n+N)),
	\end{equation*}
		where   $\mathrm{poly}(\log n)$ denotes a polylogarithmic factor bounded by $c (\log n)^r$ for some universal constants $c, r > 0$ independent of the sample sizes.
\end{restatable}

 Theorem~\ref{thm_ssl} shows that the total generative error decomposes into a structural bias $I(\widetilde{X};Y\mid \widetilde{R})$ and the statistical errors arising from the estimation of the two velocity fields. The primary advantage of the proposed factorization is that these two estimation tasks possess distinct statistical requirements and therefore leverage different data sources. Estimating the first-stage velocity field requires supervision and thus relies exclusively on the $n$ labeled samples. However, this task is confined to the lower dimension $k+q+1$. In contrast, estimating the second-stage velocity field involves a higher-dimensional reconstruction problem in dimension $d+q+1$. Because the conditional distribution $P_{X\mid R}$ does not depend on $Y$, this second stage can exploit the entire augmented dataset of size $n+N$.

 In the regime with abundant unlabeled data where $N \gg n$, the estimation error of the second stage becomes statistically negligible. Consequently, the overall convergence rate is dominated by the low-dimensional supervised component. The representation-based framework RepG thus effectively utilizes unlabeled data to mitigate the statistical burden of high-dimensional generative modeling.  This is formalized in the following corollary.

 \begin{corollary} \label{cor_SSL}
 	Suppose the conditions of Theorem \ref{thm_ssl} hold. Assume $\alpha = \beta$ and $I(X; Y \mid R) = 0$. If the unlabeled sample size satisfies $N \gtrsim n^{\frac{2\alpha+ d + k + 1}{2\alpha + q + k + 1}} - n$, then
 	\begin{equation*}
 		\mathbb{E}_{\mathcal{D}_l,\, \mathcal{D}_u,\, Y}\bigl[ D_{\mathrm{KL}}\bigl( P_{\widetilde{X} \mid Y} \,\|\, \widehat{P}^{\,\mathrm{semi}}_{X \mid Y} \bigr) \bigr]
 		\lesssim \delta(\varepsilon)+ n^{-\frac{2\alpha}{2\alpha + q + k + 1}} \mathrm{poly}(\log n).
 	\end{equation*}
 \end{corollary}

 Corollary~\ref{cor_SSL} indicates that, provided the unlabeled sample size is sufficiently large, the overall convergence rate of RepG is  determined by the first-stage estimation targeting $P_{R\mid Y}$. This  confines the statistical complexity to the lower dimension $k+q+1$ rather than the ambient dimension $d+q+1$. Abundant unlabeled data therefore effectively  mitigates the statistical cost of high-dimensional reconstruction.

\subsubsection{Fully supervised setting}
\label{sec_full}
The fully supervised setting corresponds to  a special case of the semi-supervised framework, where no unlabeled data is available ($N = 0$). In this setting, both stages are estimated  using the labeled sample $\mathcal{D}_l = \{(X_i, Y_i)\}_{i=1}^n$. We denote the resulting conditional generative distribution as $\widehat{P}^{\,\mathrm{full}}_{X \mid Y}$. Setting $N = 0$ in Theorem~\ref{thm_ssl} immediately yields the corresponding upper bound for this fully supervised estimator.

\begin{proposition}
	\label{pro_fullsuper}
	Suppose the conditions of Theorem \ref{thm_ssl} hold. Assume $\alpha = \beta$ and $I(X; Y \mid R) = 0$. Then, the expected KL divergence of the fully supervised estimator satisfies
	\begin{equation*}
		\mathbb{E}_{\mathcal{D}_l,\, Y}\bigl[ D_{\mathrm{KL}}\bigl( P_{\widetilde{X} \mid Y} \,\|\, \widehat{P}^{\,\mathrm{full}}_{X \mid Y} \bigr) \bigr]
		\lesssim \delta(\varepsilon) + n^{-\frac{2\alpha}{2\alpha + \max\{q,\, d\} + k + 1}} \mathrm{poly}(\log n).
	\end{equation*}
\end{proposition}
Proposition \ref{pro_fullsuper} shows that even in the fully supervised setting, the representation-based framework achieves a faster convergence rate than direct conditional sampling. The structural factorization   confines the effective statistical dimension to $\max\{k+q+1,\, d+k+1\}$. This directly avoids the ambient dimension $d+q+1$ that governs  direct approaches, a  limitation formalized by Theorem~\ref{thm_lowerbound}.

\subsection{Minimax lower bound for direct conditional generation}
\label{sec_lower}
To quantify the statistical advantage of the proposed framework, we establish the fundamental limits of direct conditional generation without imposing any latent structure.

We consider a direct generative approach that models the target conditional distribution $P_{X \mid Y=\by}$ as the terminal measure of a conditional SDE initialized at $X_0 \sim \mathcal{N}(0, \bI_d)$ and evolving according to
\begin{equation}\label{eq_sde_direct}
	\diff X_t = \bv_F(t, X_t; \by)\,\diff t + \sqrt{2\eta(t)}\,\diff B_t, \qquad t \in [0, 1-\varepsilon].
\end{equation}
The drift admits $\bv_F(t,\bx;\by)=\bv(t,\bx;\by)+ \eta(t)w(t) (\bv(t,\bx;\by)- (\dot b_t/b_t )\bx)$, where $w(t)$ is a known weight function. Under this formulation, the generative task  reduces to estimating a  nonparametric velocity field $\bv: [0,1-\varepsilon]\times\mathbb{R}^d\times\mathbb{R}^q\to\mathbb{R}^d$.

%

\begin{definition}[Generative conditional distribution class $\mathcal{P}_{\omega}$]
	\label{def_distribution_class}
	Let $\mathcal{P}_{\omega}$ denote the class of conditional path measures $\mathbf{P}_{X \mid Y}$ over $[0, 1-\varepsilon]$ satisfying the following two conditions.
	\begin{enumerate}
		\item[(i)] For any given $\by$, the path measure $\mathbf{P}_{X \mid Y=\by}$ is generated by the SDE in \eqref{eq_sde_direct}.
		\item[(ii)] The corresponding velocity field $\bv=(v_1,\dots, v_d)^{\top}$ satisfies the localized $\omega$-order H{\"o}lder smoothness condition, in the sense of Assumption~\ref{ass_smoothness} with smoothness parameter $\omega$.
	\end{enumerate}
\end{definition}

\begin{restatable}{theorem}{thmlower}
	\label{thm_lowerbound}
	Consider any conditional path measure estimator $\widehat{\mathbf{P}}_{X \mid Y}$ generated via the SDE \eqref{eq_sde_direct}, driven by an arbitrary conditional velocity field estimator $\widehat{\bv}$ constructed from the labeled dataset $\mathcal{D}_l$ of size $n$. For any valid marginal distribution $P_Y$, the minimax expected KL divergence satisfies
	\begin{equation}\label{eq_lower_bound}
		\inf_{\widehat{\mathbf{P}}_{X \mid Y}} \sup_{\mathbf{P}_{X \mid Y} \in \mathcal{P}_{\omega}} \mathbb{E}_{\mathcal{D}_l} \mathbb{E}_{Y}\bigl[D_{\mathrm{KL}}\bigl( \mathbf{P}_{X \mid Y} \,\|\, \widehat{\mathbf{P}}_{X \mid Y} \bigr)\bigr] \gtrsim n^{-\frac{2\omega}{2\omega + d + q + 1}}.
	\end{equation}
\end{restatable}

Theorem \ref{thm_lowerbound} shows the statistical limitation of direct conditional generation. The effective dimension governing the estimation error is the full ambient dimension $d+q+1$. This lower bound characterizes the fundamental statistical cost of learning the unconstrained velocity field. While the upper bound in Proposition~\ref{pro_fullsuper} evaluates the terminal generative error, both results
share the same effective dimension structure, and the comparison reflects the gain from imposing the representation factorization. Partitioning the generative process avoids the combined $d+q$ dependence, yielding an upper bound governed by the reduced effective dimension $\max\{q, d\} + k + 1$.

\begin{remark}
Theorem~\ref{thm_lowerbound} distinguishes SDE-based generative modeling from classical conditional density estimation. Standard static density estimation over $\mathbb{R}^d \times \mathbb{R}^q$ achieves the minimax optimal rate of $n^{- 2\omega / (2\omega + d + q)}$ \citep{Tsybakov2009}. In contrast, continuous-time generative models require estimating a time-dependent velocity (or drift) field. Because the divergence between path measures is strictly governed by the time-integrated estimation error of this drift field, time acts as an active regression coordinate. This mathematically imposes a $+1$ dimensionality penalty on the minimax rate, as established in \eqref{eq_lower_bound}. Consequently, sampling via continuous dynamics incurs a strictly higher statistical sample complexity than classical static density estimation under equivalent smoothness conditions.
\end{remark}

\section{Numerical experiments}\label{sec_num}
In this section, we evaluate the finite-sample performance of the proposed representation-based generative method RepG. We first present simulations that show its capacity to learn diverse data structures, and then turn to an application on the MNIST dataset.

\subsection{Simulation}
We conduct simulation studies to evaluate the conditional generative performance of RepG under both discrete and continuous settings.  To construct sufficient representations, we apply the deep dimension reduction (DDR) framework of \citet{HuangSRL2024}. This method learns a mapping $R:\mathbb{R}^d \to \mathbb{R}^k$ that forces $R(X)$ to follow a standard Gaussian distribution while satisfying the conditional independence $X \inde Y \mid R(X)$. This ensures $R(X)$ retains all predictive information about $Y$ to facilitate subsequent conditional generative modeling. We optimize this mapping efficiently by maximizing the empirical distance covariance \citep{Szekelye2007} regularized by an energy distance penalty.

\subsubsection{Discrete conditioning}
For the case where $Y$ is discrete, we generate $n=1500$ observations per class across four synthetic datasets with distinct geometric structures: (1) Two moons: two interleaving crescent-shaped clusters with nonconvex support; (2) Concentric circles: two radially arranged classes with inner radius $r_0=0.5$ and outer radius $r_1=1.5$; (3) 2D Gaussian mixture: four classes at the corners of a square, each distributed as $\mathcal{N}(\mu_c, 0.3^2 I_2)$; and (4) 3D Gaussian mixture: six isotropic components with variance $0.3^2$ and means at $(0,0,0)$, $(2,0,0)$, $(0,2,0)$, $(0,0,2)$, $(2,2,0)$, and $(2,0,2)$.

\begin{figure}[htbp]
	\centering
	\setlength{\tabcolsep}{1.5pt}
	\renewcommand{\arraystretch}{1.2}
	
	\begin{tabular}{c c c c c}
		&  {\footnotesize Two Moons} &  {\footnotesize Circle} &  {\footnotesize 2D Gaussian} &  {\footnotesize 3D Gaussian} \\
		
		\raisebox{2.3em}{\rotatebox[origin=c]{90}{ {\scriptsize Real $X|Y$}}} &
		\includegraphics[width=0.12\linewidth, height=0.12\linewidth]{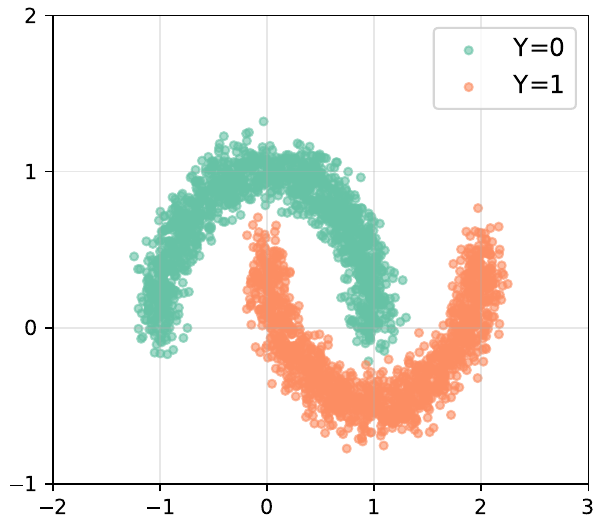} &
		\includegraphics[width=0.12\linewidth, height=0.12\linewidth]{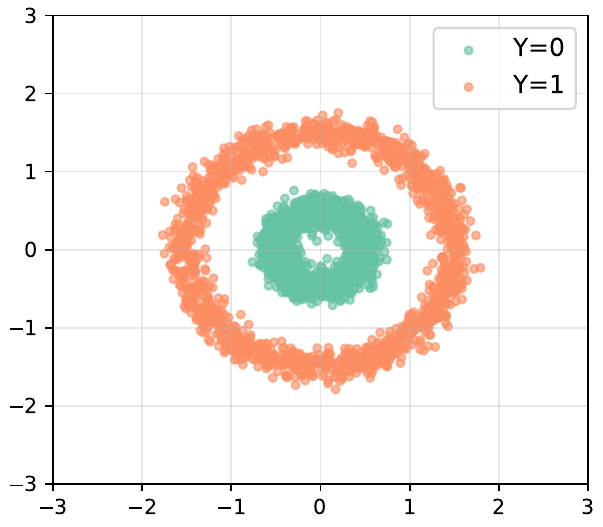} &
		\includegraphics[width=0.12\linewidth, height=0.12\linewidth]{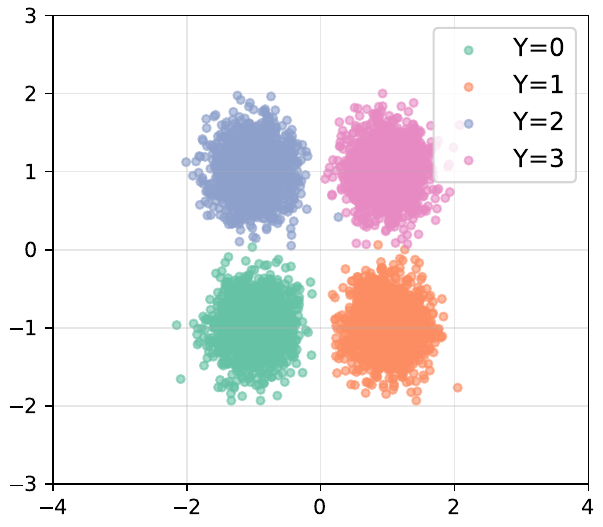} &
		\includegraphics[width=0.12\linewidth, height=0.12\linewidth]{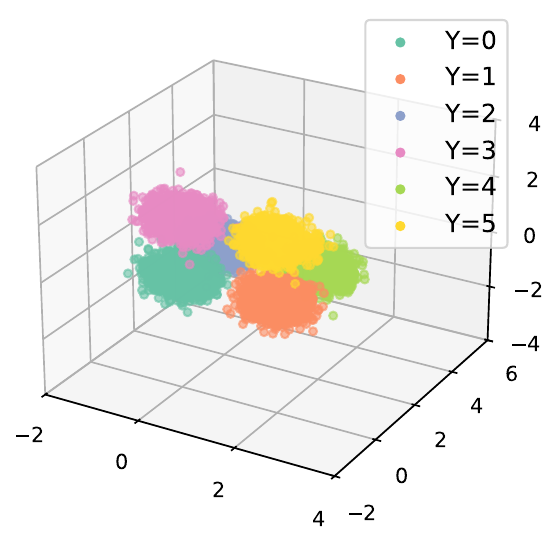} \\
		
		\raisebox{2.4em}{\rotatebox[origin=c]{90}{{\scriptsize Generated $R|Y$}}}  &
		\includegraphics[width=0.12\linewidth, height=0.12\linewidth]{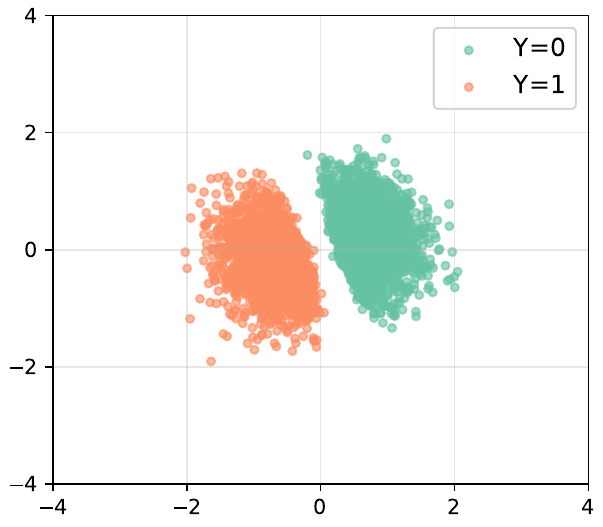} &
		\includegraphics[width=0.12\linewidth, height=0.12\linewidth]{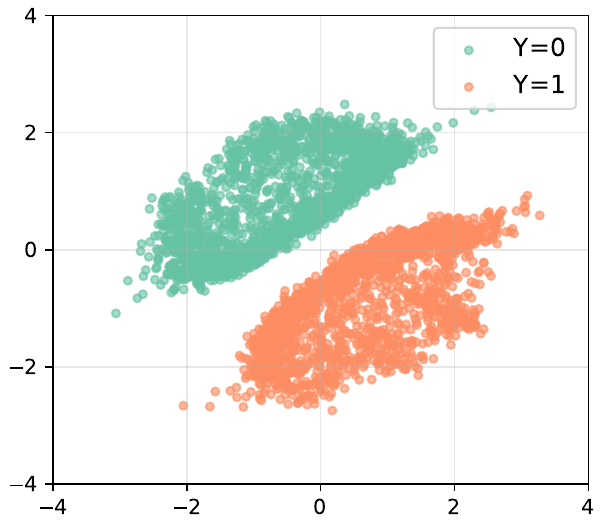} &
		\includegraphics[width=0.12\linewidth, height=0.12\linewidth]{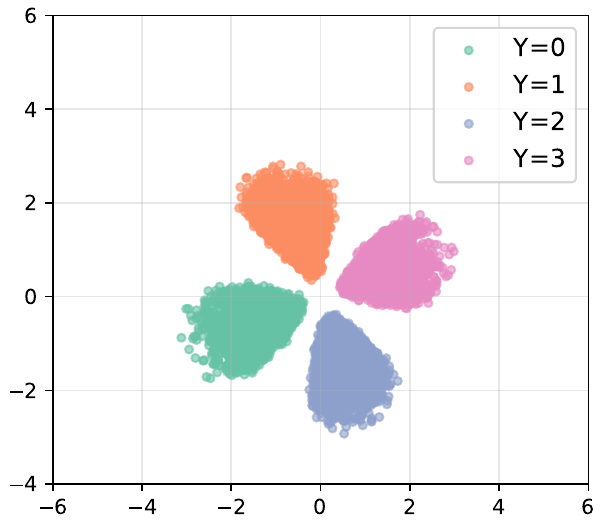} &
\includegraphics[width=0.12\linewidth, height=0.12\linewidth]{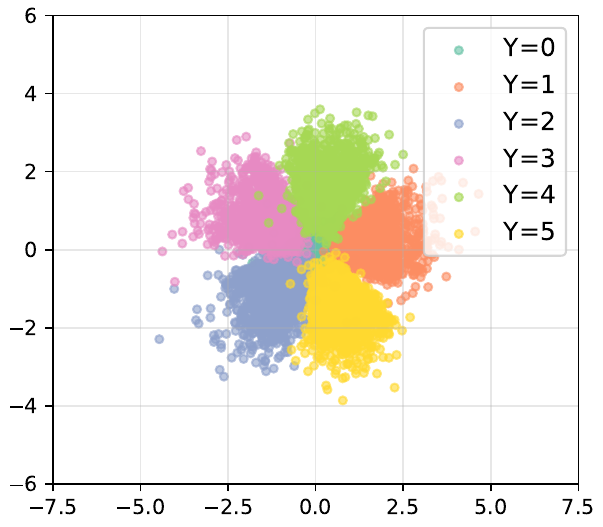} \\	
		\raisebox{2.2em}{\rotatebox[origin=c]{90}{{\scriptsize Generated $X|Y$}}} &
		\includegraphics[width=0.12\linewidth, height=0.12\linewidth]{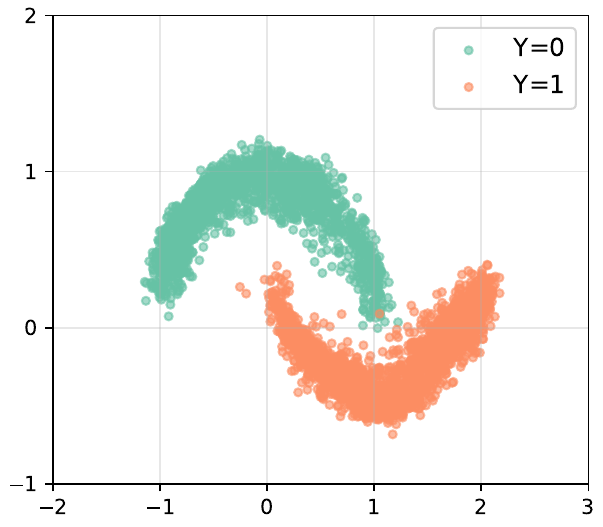} &
		\includegraphics[width=0.12\linewidth, height=0.12\linewidth]{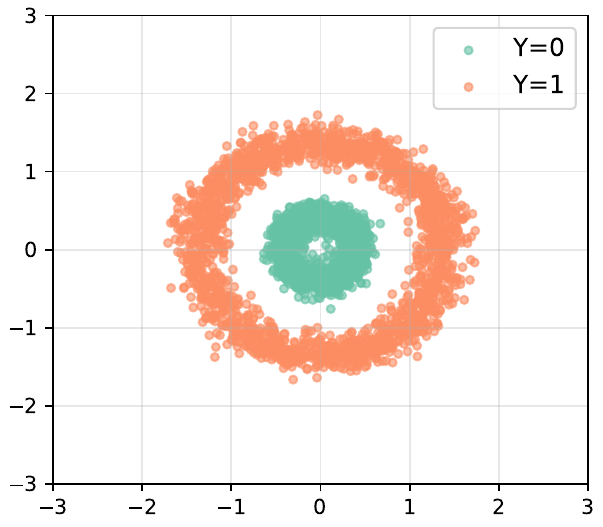} &
		\includegraphics[width=0.12\linewidth, height=0.12\linewidth]{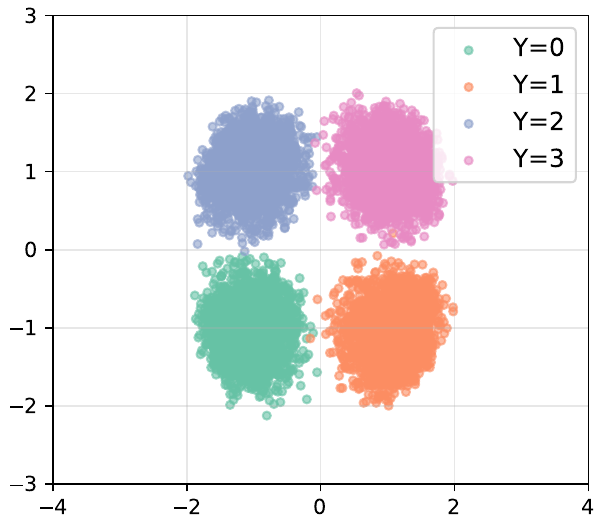} &
		\includegraphics[width=0.12\linewidth, height=0.12\linewidth]{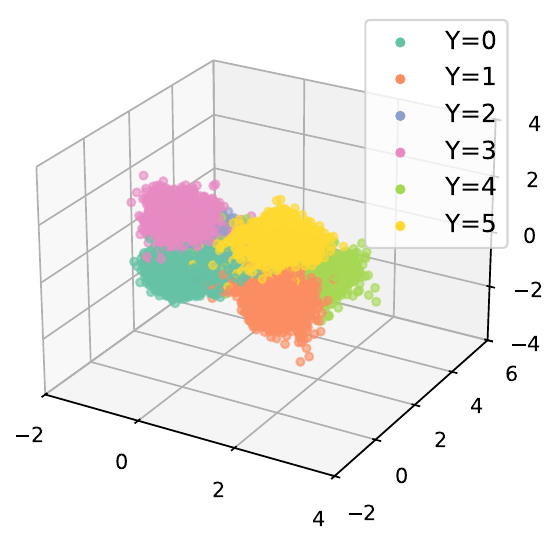} \\
	\end{tabular}
	\caption{Generative performance of RepG across four distinct scenarios. The top row illustrates the real data  $X|Y$. The middle row displays the generated conditional representations  $R|Y$. The bottom row shows the final generated samples $X|Y$.}
	\label{fig_class_exp}
\end{figure}

Figure~\ref{fig_class_exp} illustrates RepG's generative capabilities across four distinct scenarios. The DDR mapping learns a standard Gaussian latent space that preserves strict class separability for complex, nonlinear data, a structure that is visually apparent in the generated conditional representations (middle row).
By mapping these generated latent samples back to the original domain, RepG  reconstructs the complex geometries of the datasets without suffering from mode collapse or structural degradation (bottom row).

\subsubsection{Continuous conditioning}
For the case where $Y$ is continuous, we consider three scenarios, generating $n = 3000$ training pairs and
$n_{\mathrm{test}} = 500$ test pairs $(X, Y)$, where $X \in \mathbb{R}^{10}$ and $Y \in \mathbb{R}$.
Each $X=(X^{(1)},\dots, X^{(10)})^{\top}$ is obtained by embedding a scalar latent signal $X_0 \in \mathbb{R}$ into $\mathbb{R}^{10}$ via
$
X^{(1)} = X_0,
X^{(j)} = f_j(X_0) + \varepsilon_j,  j = 2,\ldots,10, \quad \varepsilon_j \overset{\mathrm{iid}}{\sim} \mathcal{N}(0,\,0.05^2),
$
where the transformation maps $f_j$ cycle through four nonlinear families indexed by $j \bmod 4$: linear scaling, hyperbolic tangent, signed power, and cubic.
Every coordinate carries information about $X_0$, meaning the intrinsic dimension of $X$ is strictly one. The conditional distribution of the latent signal $X_0$ is specified for each scenario as follows.
\begin{enumerate}[label=(\arabic*), leftmargin=2em, itemsep=4pt]
	\item {Heteroscedastic:}\;
$X_0 | Y = y \sim \mathcal{N}\! (2\sin(y) + 0.5y, (0.2 + 0.4\lvert y\rvert)^2 ),$ with $Y \sim \mathrm{Unif}(-4, 4)$.

	\item {Skewed:}\;
$X_0 |  Y = y   \sim 2\sin(y) +  (0.5 + 0.2\lvert y\rvert )  (\xi - e^{\tau^2/2} ),$ where $ \xi \sim \mathrm{LogNormal}(0, \tau^2), \tau = 0.8,$ and $Y \sim \mathrm{Unif}(-3, 3)$.
	\item {Crossing:}\;
$X_0 | Y = y  \sim \tfrac{1}{2}  \mathcal{N}(y,0.04) + \tfrac{1}{2} \mathcal{N}(-y,0.04)
$, with $Y \sim \mathrm{Unif}(-3, 3)$.
\end{enumerate}

Within the framework of conditional distribution estimation, RepG is evaluated against three established competing methods:  Gaussian process regression (GPR) \citep{rasmussen2006gaussian}, a mixture density network (MDN, $K = 5$)
\citep{bishop1995neural}, and a quantile random forest (QRF) \citep{meinshausen2006quantile}.  These methods predict each coordinate of $X$ independently, either by design (GPR, QRF) or due to the practical instability of fitting high-dimensional joint mixtures (MDN).  To assess generative performance, we first draw synthetic multivariate samples $\hat{X} \in \mathbb{R}^{10}$ from each trained model conditioned on the test observations of $Y$. We then evaluate these generated samples across two distinct criteria. First, to assess the recovery of the multivariate dependence structure, we measure the joint distributional accuracy using the energy score (ES) computed on the full $\mathbb{R}^{10}$ space \citep{gneiting2007strictly}. Second, we evaluate marginal interval efficiency using local conformal prediction \citep{lei2018distribution} applied specifically to the latent signal coordinate $X^{(1)}$. We calibrate this marginal prediction interval to ensure all methods achieve a nominal $95\%$ coverage probability ($\mathrm{PICP}_{\mathrm{cal}}$). Subsequently, we compare the mean prediction interval width ($\mathrm{MPIW}_{\mathrm{cal}}$), where a narrower calibrated interval indicates a more precise and informative conditional marginal distribution.

\begin{figure}[t]
	\centering
	\setlength{\tabcolsep}{1.5pt}
	\renewcommand{\arraystretch}{1}
	\begin{tabular}{ccc}
		{Heteroscedastic} &  {Skewed} & {Crossing} \\
		\includegraphics[width=0.22\linewidth, height=0.2\linewidth]{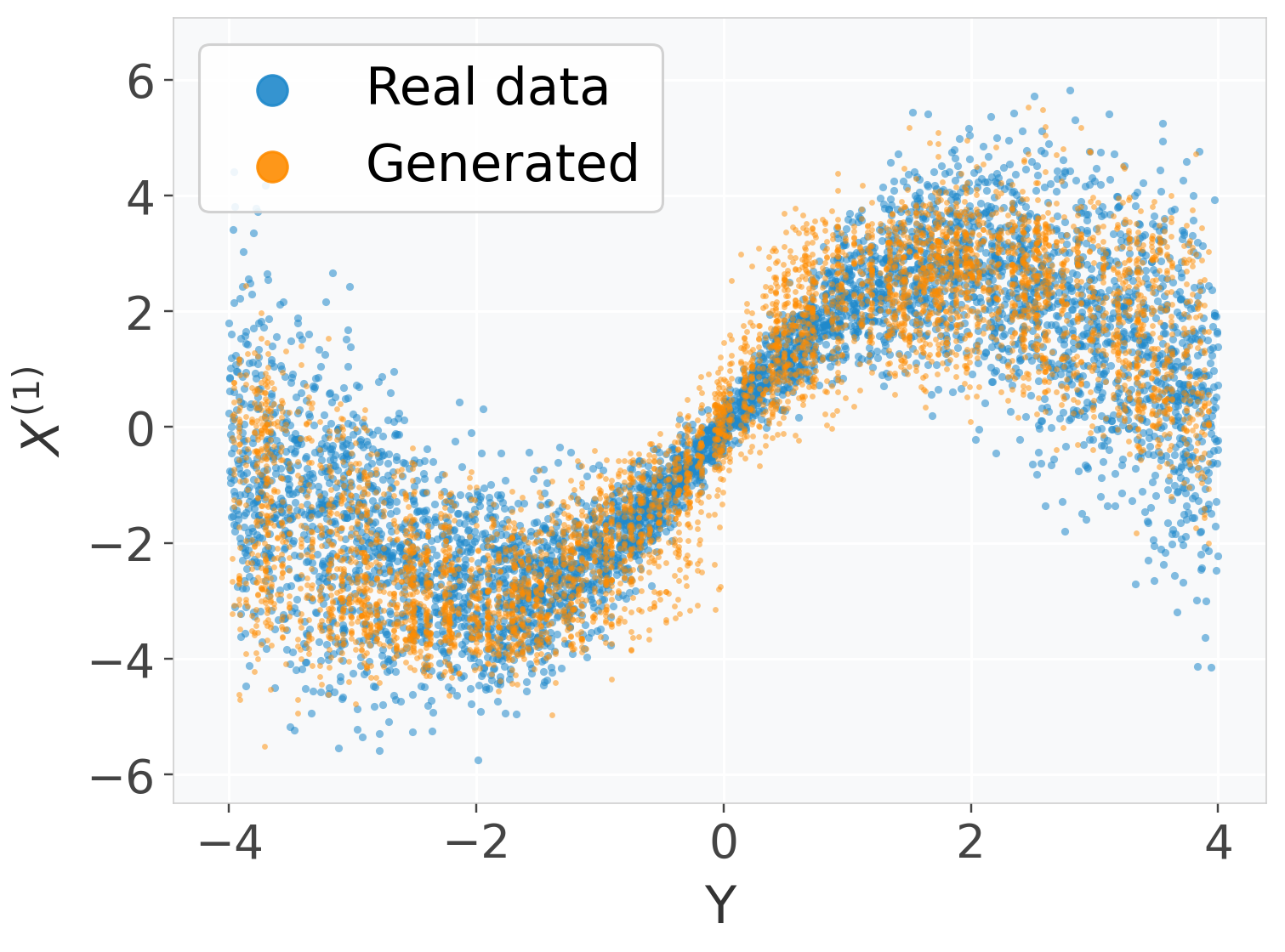} &
		\includegraphics[width=0.22\linewidth, height=0.2\linewidth]{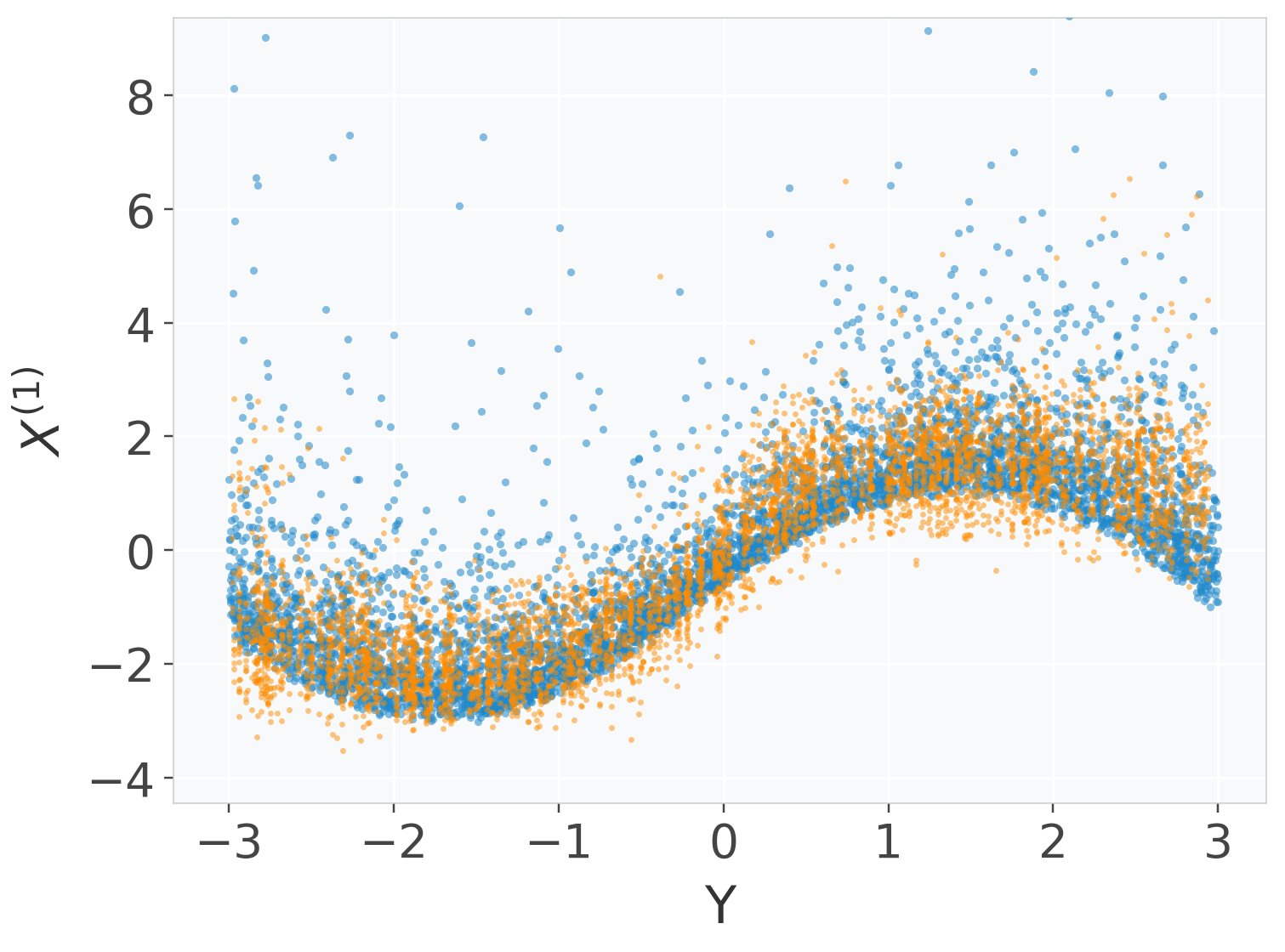} &
		\includegraphics[width=0.22\linewidth, height=0.2\linewidth]{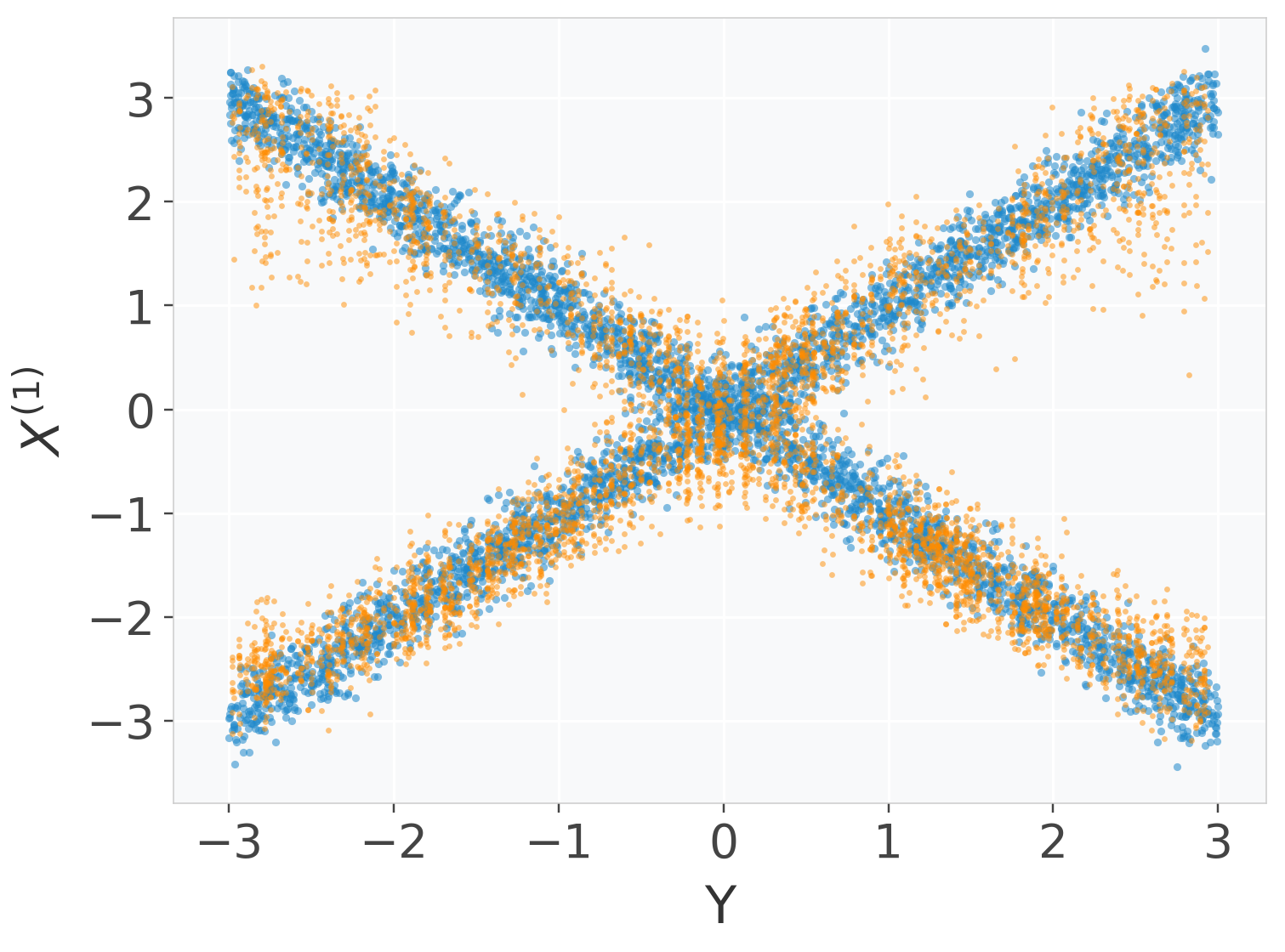} \\
	\end{tabular}
	\caption{Real data (blue) and RepG generated samples (orange) across three distinct scenarios. Each panel displays $X^{(1)}$ samples conditioned on $Y$.}
	\label{fig_reg}
\end{figure}

\begin{table}[H]
	\centering
	\small
	\caption{
		Comparison of methods for conditional distribution estimation in $\mathbb{R}^{10}$. Methods marked (D) fit each dimension independently.  For ES and $\mathrm{MPIW}_{\mathrm{cal}}$, lower is better, with the best result in each dataset bolded.
		The $\mathrm{PICP}_{\mathrm{cal}}$ is reported merely to confirm valid nominal coverage (target $0.95$) prior to comparing interval efficiency.}
	\label{tab:regression_results}
	\renewcommand{\arraystretch}{1.0}
	\setlength{\tabcolsep}{6pt}
	\begin{tabular}{llccc}
		\toprule
		\multirow{2}{*}{\textbf{Dataset}}
		& \multirow{2}{*}{\textbf{Method}}
		& \multicolumn{1}{c}{\textit{Joint acc.}}
		& \multicolumn{2}{c}{\textit{Interval eff.  }} \\
		 \cmidrule(lr){3-3}\cmidrule(lr){4-5}
		&& ES$\downarrow$
		& $\mathrm{PICP}_{\mathrm{cal}}$
		& $\mathrm{MPIW}_{\mathrm{cal}}$$\downarrow$ \\
		\midrule
		
		\multirow{4}{*}{Heteroscedastic}
		& GPR\,(D)  &  2.903 & 0.964 & 4.720 \\
		& MDN\,(D)  &  2.649 & 0.956 & \textbf{4.519} \\
		& QRF\,(D)  &  2.723 &  {0.948} & 4.761 \\
		& RepG      &   \textbf{2.615} & 0.960 & 4.836 \\
		\midrule
		
		\multirow{4}{*}{Skewed}
		& GPR\,(D)  &  5.139 & 0.960 & 5.756 \\
		& MDN\,(D)  &  {2.189} & 0.964 & \textbf{5.299} \\
		& QRF\,(D)  &  2.303 & 0.956 & 5.750 \\
		& RepG      &  \textbf {2.188 }&  {0.964} & 5.786 \\
		\midrule
		
		\multirow{4}{*}{Crossing}
		& GPR\,(D)  &  3.684 & 0.968 & 5.231 \\
		& MDN\,(D)  &  3.458 & 0.956 & \textbf{3.693} \\
		& QRF\,(D)  &   3.529 & 0.964 & 3.739 \\
		& RepG      &  \textbf{2.849} &  {0.952} & 3.778 \\
		\bottomrule
	\end{tabular}
\end{table}

  Figure~\ref{fig_reg} plots conditional samples of the first coordinate $X^{(1)}$ generated by RepG against real observations across three synthetic scenarios. The visualizations show that the proposed framework faithfully recovers these distinct conditional geometries.

As shown in Table~\ref{tab:regression_results}, RepG achieves high joint distributional accuracy across the three datasets while maintaining competitive marginal performance. It attains the best ES in all the scenarios, including the complex Crossing case, showing superior recovery of cross-dimensional dependence. Although independent per-dimension methods yield slightly narrower calibrated intervals ($\mathrm{MPIW}_{\mathrm{cal}}$), they do not explicitly model the multivariate dependence structure.  By preserving joint geometry, RepG substantially improves ES at the cost of only modestly wider marginal intervals. Finally, conformal recalibration ensures all methods achieve a $\mathrm{PICP}_{\mathrm{cal}}$ within 1.8 percentage points of the $95\%$ nominal level, providing valid coverage for the interval efficiency comparisons.

\subsection{Real data application}
 We evaluate RepG on the MNIST dataset \citep{Lecun1998},
  comprising $28{\times}28$ grayscale images of handwritten digits ($0$--$9$), accessed via \texttt{torchvision}. We subsample $10000$ training and $5000$ test images, scaling pixel values to $[-1,1]$.  We compare the proposed RepG against a baseline that directly models the conditional distribution $P_{X\mid Y}$. To  assess performance in the semi-supervised setting, the labeled date size is fixed at $n_\ell \in \{1000, 3000\}$, while the unlabeled date size varies over $n_u \in \{0, 1000, 2000, 3000, 4000, 5000\}$.

 We parameterize the DDR mapping via a convolutional neural network that projects images into a latent space $\mathbb{R}^{32}$. To ensure strict independence, the DDR model is pre-trained on an auxiliary set of $5000$ samples, completely disjoint from the generative training phase. For the conditional generative stages, we employ a U-Net architecture with a linear interpolation schedule. The SDE discretization utilizes $100$ steps per stage with $\eta=0.2$. To evaluate generative quality, we synthesize $200$ images per class and compute the classification accuracy using a strong CNN classifier pre-trained on the original MNIST dataset. This metric assesses the fidelity of the synthesized samples by checking if they can be correctly identified as their intended classes.

\begin{figure}[H]
 \centering
 \begin{tabular}{cc}
	\includegraphics[width=0.32\linewidth]{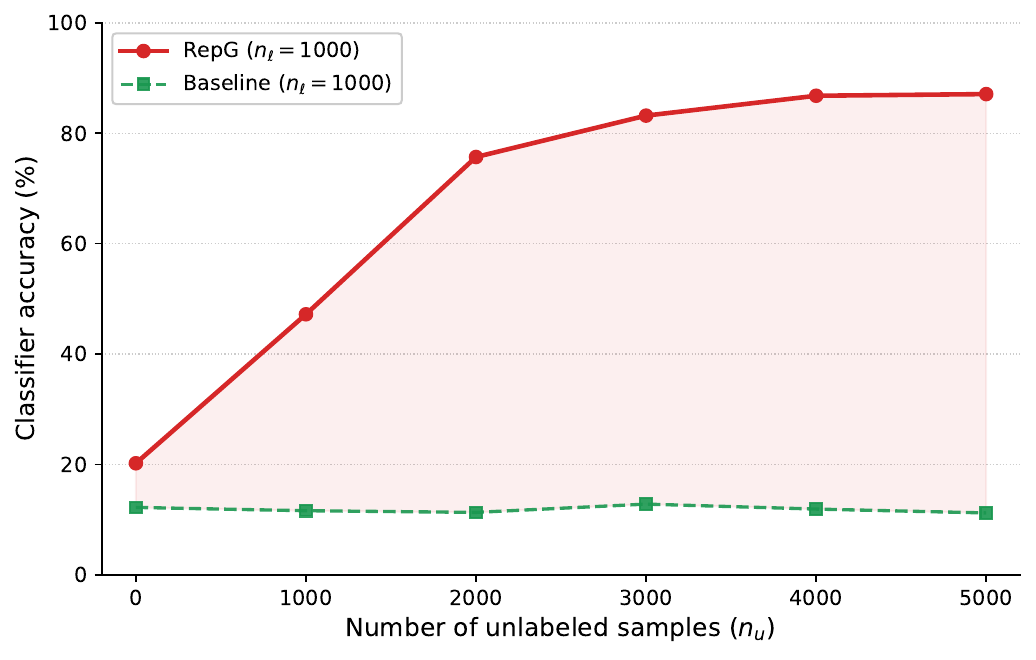}&
\includegraphics[width=0.32\linewidth]{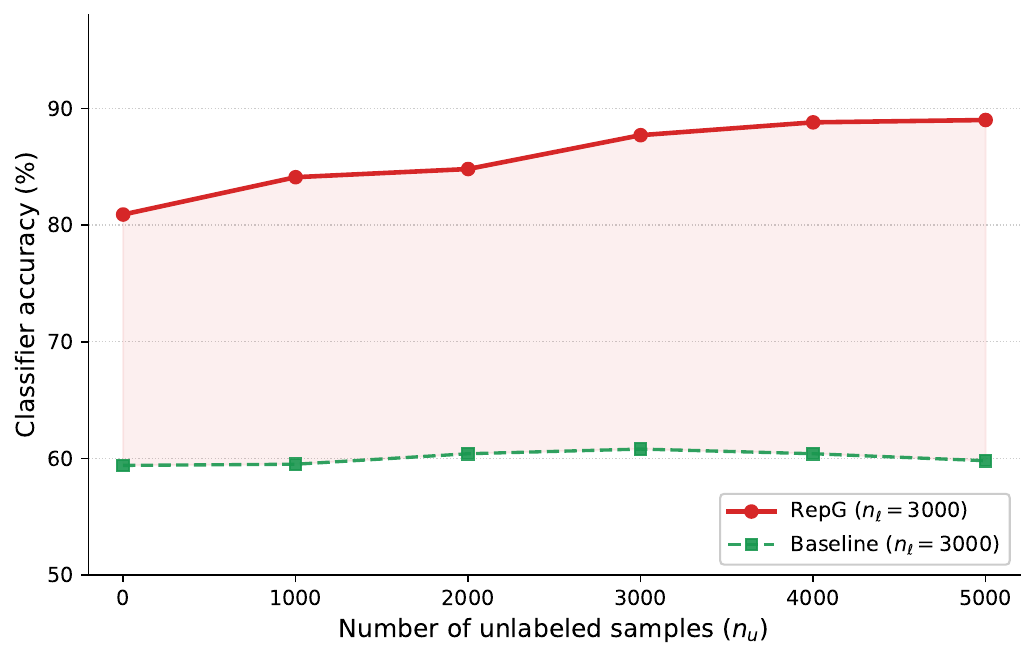}\\
(a)\ $n_{\ell}=1,000$  & (b)\ $n_{\ell}=3,000$
\end{tabular}
\caption{Comparison of classifier accuracy between  RepG (red)  and baseline (green) against $n_u$ from $0, 1,000, 2,000$ to $5,000.$ (a) left panel: $n_{\ell}=1,000$; (b) right panel: $n_{\ell}=3,000.$}
\label{fig_accu1000}
\end{figure}

\bigskip
Figures~\ref{fig_accu1000}(a)
plots classifier accuracy against the number of unlabeled set ($n_u$) for a fixed labeled set sizes of  $n_\ell = 1,000.$   Since the baseline relies only on labeled data, its performance remains invariant with respect to $n_u$, resulting in an essentially flat curve.  RepG consistently outperforms this baseline across all values of $n_u$.
In the fully supervised setting ($n_u = 0$), RepG achieves 20.2\% accuracy compared to 12.0\% for the baseline. These absolute accuracies are  relatively low, because $n_\ell = 1000$ provides only 100 labeled examples per class across the ten digit classes. The primary focus, however, is the relative improvement of RepG and its monotonic accuracy gains as $n_u$ increases. Specifically, the accuracy of RepG grows  from 20.2\% and stabilizes near $87\%$ as $n_u$ approaches $5000$.   This trajectory demonstrates the statistical advantage of the representation-based framework in data-scarce settings.
Figure~\ref{fig_accu1000} (b)  plots classifier accuracy with $n_\ell=3,000.$ With more labeled data, the baseline achieves a higher initial accuracy than in the $n_\ell = 1,000$ setting, yet RepG maintains a consistent performance advantage throughout. The accuracy of RepG increases monotonically with $n_u$ and stabilizes near 90\%.

\begin{figure}[H]
 \centering
 \begin{tabular}{cc}
	\includegraphics[width=0.28\linewidth]{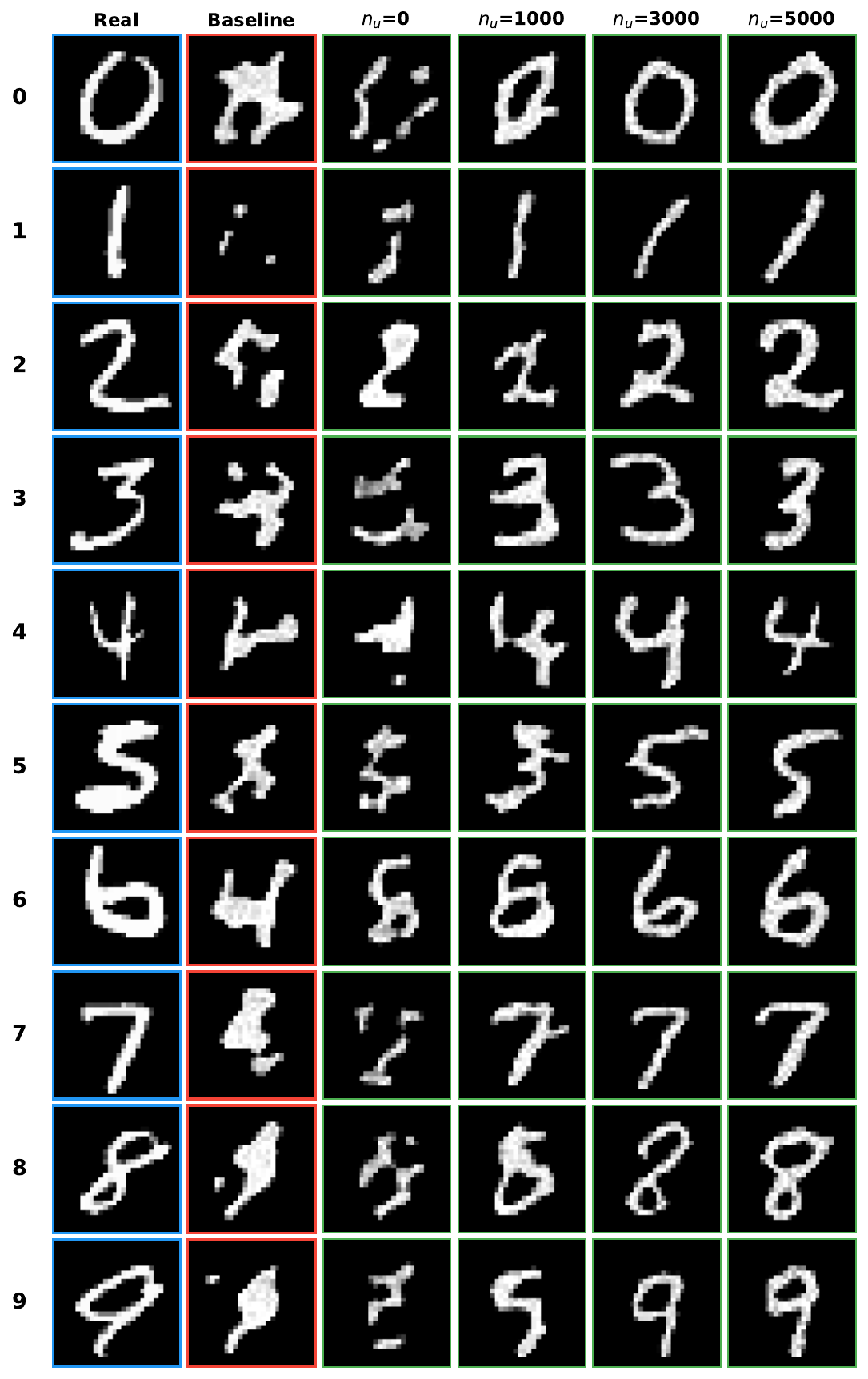}&
\includegraphics[width=0.28\linewidth]{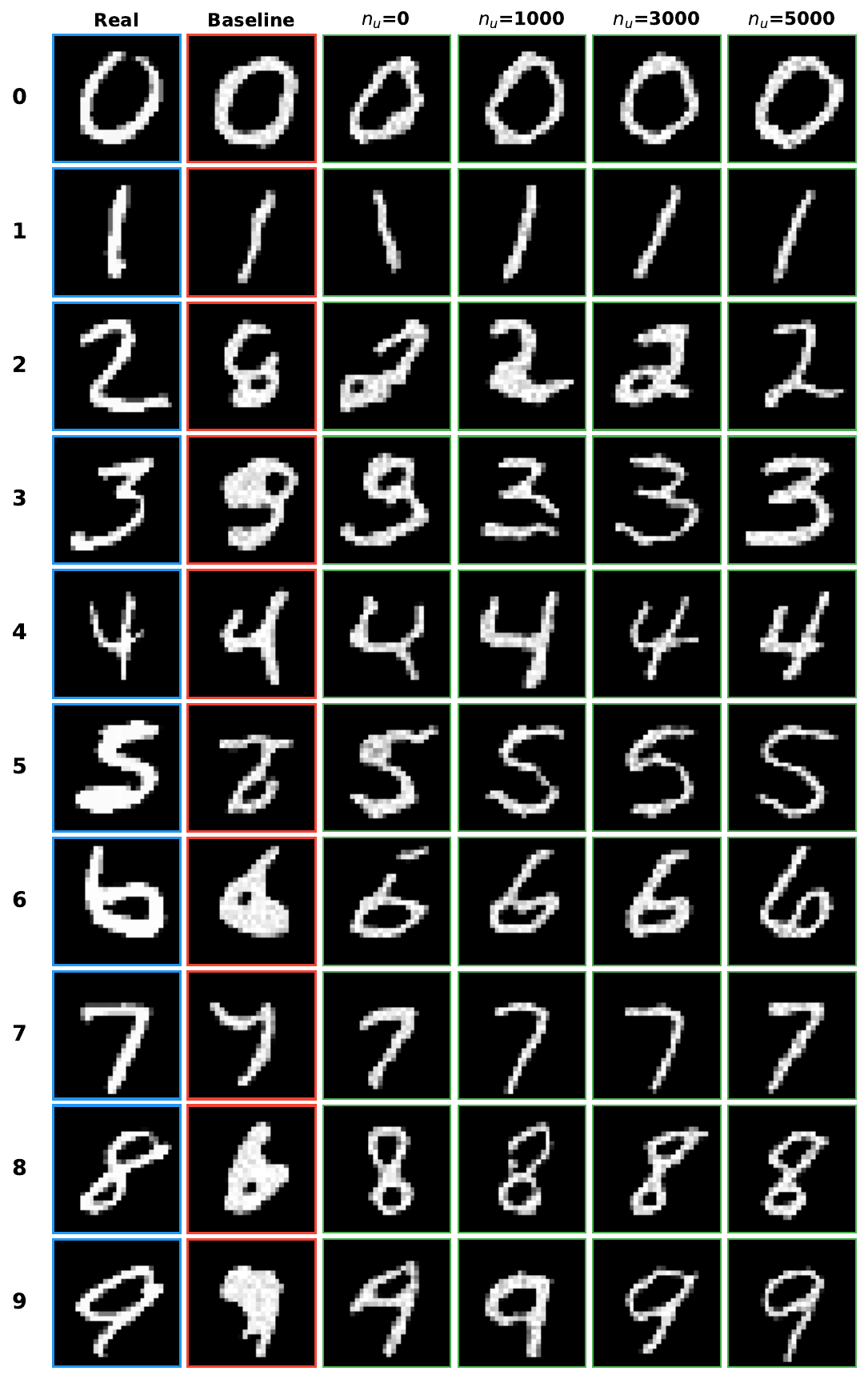}\\
$(a)\ n_{\ell}=1,000$  &(b)\  $n_{\ell}=3,000$
\end{tabular}
	\caption{Comparison of generation quality: (a) left panel $n_\ell=1,000$; (b) right panel $n_\ell=3,000.$
In each panel, the leftmost columns display the real data and samples generated by the baseline model. The subsequent columns illustrate the output of the proposed RepG method as the number of unlabeled samples $n_u$ increases from $0, 1,000, 3,000$ to $5000$.}
\label{Fig_com1000}
\end{figure}

Figure~\ref{Fig_com1000} (a)  provides visual confirmation of the generation results with $n_{\ell}=1,000$
for the MNIST data.  As unlabeled sample size $n_u$ increases, the generated digits transition from fragmented artifacts to coherent structures.
Figure~\ref{Fig_com1000} (b) presents
visualizations for larger labeled sets ($n_\ell = 3000$) , which shows similar structural improvements as $n_u$ increases.

\section{Conclusion}\label{sec_con}

In this paper, we proposed RepG, a semi-supervised framework for conditional generative modeling that combines conditional stochastic interpolation with low-dimensional latent representations. By approximating the target distribution $P_{X \mid Y}$ through the structural factorization $Y \rightarrow R \rightarrow X$, RepG decomposed the difficult high-dimensional conditional generation problem into two distinct, manageable tasks: a label-dependent latent sampling stage and a high-dimensional reconstruction stage.  We established a KL risk decomposition, identifying the conditional mutual information $I(X;Y \mid R)$ as the fundamental structural approximation bias. This explicitly links the sufficiency of the chosen representation to the total generative error.

This representation-based factorization in RepG is particularly advantageous in the semi-supervised regime. Limited labeled observations (size $n$) are utilized efficiently to characterize the low-dimensional conditional dependence between $R$ and $Y$. Concurrently, the high-dimensional reconstruction of $X$ from $R$ via stochastic interpolation is learned by leveraging the pooled information from abundant unlabeled data (size $N$). Through a non-asymptotic theoretical analysis of deep neural network estimators, we proved that this approach achieves a strict statistical rate separation. Complemented by a minimax lower bound, our theory shows that in the practical regime where $N \gg n$, the generative error of RepG is governed primarily by the estimation of the latent conditional distribution. Because this rate is governed by the reduced dimensions $k$ and $q$ rather than the ambient dimension $d$, our proposed method  bypasses the curse of dimensionality inherent in direct conditional generative approaches.

Finally, numerical experiments across synthetic and real-world datasets corroborate our theoretical findings. The results validate RepG's ability to accurately capture complex, high-dimensional data geometries while demonstrating superior statistical sample efficiency in data-scarce regimes. Future work may explore the joint, end-to-end learning of the representation mapping $R$ alongside the stochastic interpolation velocity fields to further optimize the trade-off between structural bias and estimation variance.

\medskip\noindent
\textbf{Data Availability Statement}
The MNIST data are available at the following URL: http://yann.lecun.com/exdb/mnist/ .

\spacingset{1}

\bibliographystyle{apalike} 
\bibliography{RepG0718}

\end{document}